\documentclass[]{bytedance_seed}
\usepackage[utf8]{inputenc}
\usepackage[toc,page,header]{appendix}

\usepackage{minitoc}

\title{Solving Formal Math Problems by Decomposition and\\Iterative Reflection}

\author[1 \dagger]{Yichi Zhou}
\author[1]{Jianqiu Zhao}
\author[1]{Yongxin Zhang}
\author[2, *]{Bohan Wang}
\author[1]{Siran Wang}
\author[1]{Luoxin Chen}
\author[1]{Jiahui Wang}
\author[1]{Haowei Chen}
\author[1]{Allan Jie}
\author[1]{Xinbo Zhang}
\author[1]{Haocheng Wang}
\author[1]{Luong Trung}
\author[1]{Rong Ye}
\author[1]{Phan Nhat Hoang}
\author[3]{Huishuai Zhang}
\author[1, \dagger]{Peng Sun}
\author[1, \dagger]{Hang Li}

\affiliation[1]{ByteDance Seed}
\affiliation[2]{University of science and technology of China}
\affiliation[3]{Peking University}
\contribution[*]{Work done at ByteDance Seed}
\contribution[\dagger]{Corresponding authors}

\abstract{
General-purpose Large Language Models (LLMs) have achieved remarkable success in intelligence, performing comparably to human experts on complex reasoning tasks such as coding and mathematical reasoning. However, generating formal proofs in specialized languages like Lean 4 remains a significant challenge for these models, limiting their application in complex theorem proving and automated verification. Current approaches typically require specializing models through fine-tuning on dedicated formal corpora, incurring high costs for data collection and training. In this work, we introduce \textbf{Delta Prover}, an agent-based framework that  orchestrates the interaction between a general-purpose LLM and the Lean 4 proof environment. Delta Prover leverages the reflection and reasoning capabilities of general-purpose LLMs to interactively construct formal proofs in Lean 4, circumventing the need for model specialization. At its core, the agent integrates two novel, interdependent components: an algorithmic framework for reflective decomposition and iterative proof repair, and a custom Domain-Specific Language (DSL) built upon Lean 4 for streamlined subproblem management. \textbf{Delta Prover achieves a state-of-the-art 95.9\% success rate on the miniF2F-test benchmark, surpassing all existing approaches, including those requiring model specialization.} Furthermore, Delta Prover exhibits a significantly stronger test-time scaling law compared to standard Best-of-N proof strategies. Crucially, our findings demonstrate that general-purpose LLMs, when guided by an effective agentic structure, possess substantial untapped theorem-proving capabilities. This presents a computationally efficient alternative to specialized models for robust automated reasoning in formal environments.
}

\date{\today}
\correspondence{Yichi Zhou at \email{zhouyichi.123@bytedance.com},
Peng Sun at \email{wanhesong@bytedance.com}, Hang Li at \email{lihang.lh@bytedance.com}}

\checkdata[Project Page]{\url{https://github.com/ByteDance-Seed/lean4-agent}}

\begin{document}
\maketitle

%不需要目录就注释掉 注意目录不要和第一页放在一块 要有\newpage
%\newpage
%\tableofcontents
%\newpage

\begin{figure}[htbp]
    \centering
    \includegraphics[width=0.9\textwidth]{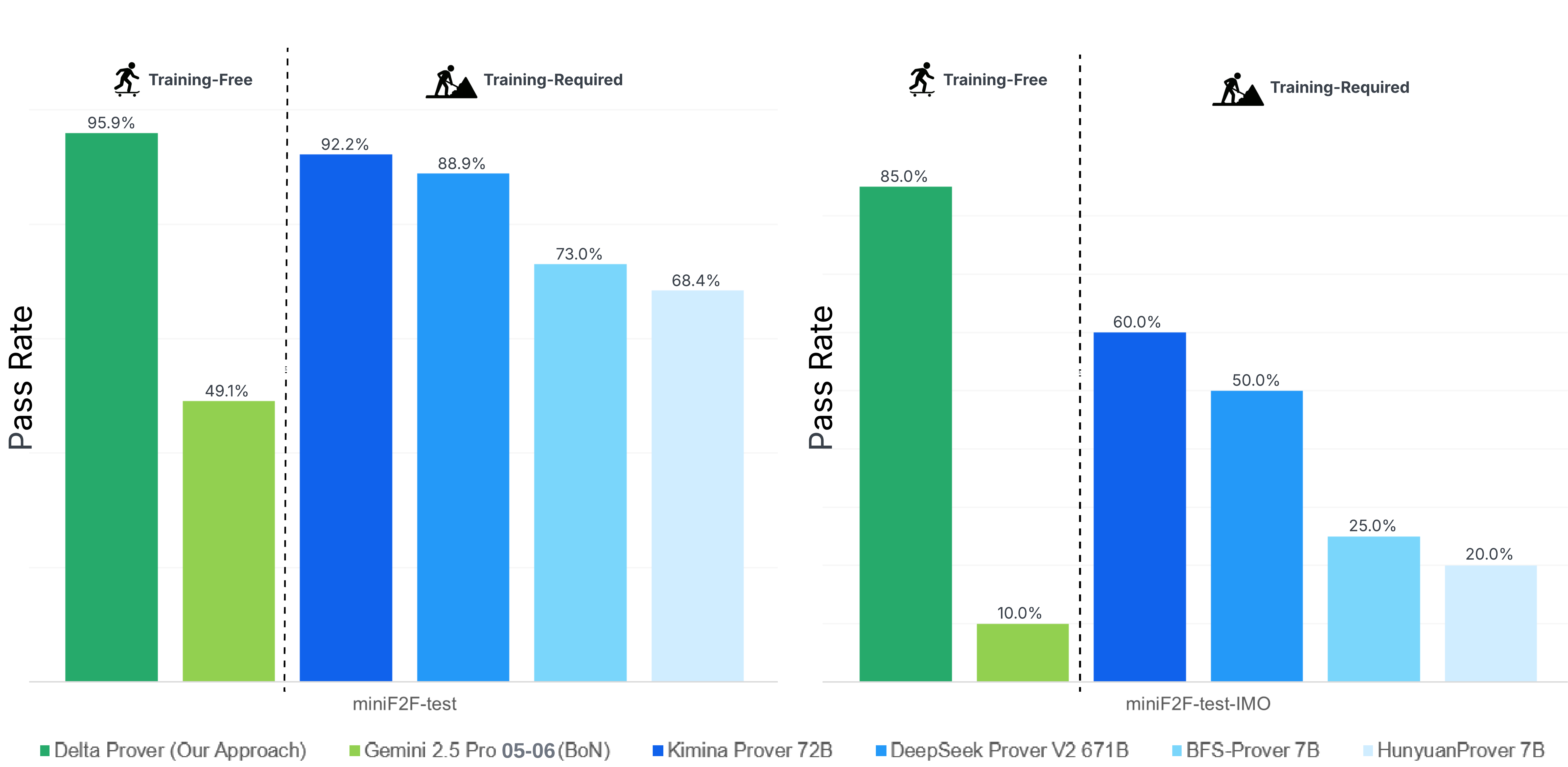}  % 替换为你的图片文件名
    \caption{Performance comparison between Delta Prover (our approach) and existing works. Our approach achieves the state-of-the-art performance over the miniF2F-test and the miniF2F-test-IMO benchmarks without additional data collection and fine-tuning, using the stock Gemini 2.5 Pro 05-06 model as a backbone. "Training-Free" and "Training-Required" stand for whether the model needs to be additionally fine-tuned over formal datasets.}
    \label{fig:comp_result}
\end{figure}

\section{Introduction}

Recent large language models (LLMs) have demonstrated remarkable progress in intellectually demanding tasks, such as solving math word problems \cite{jaech2024openai}, code generation \cite{el2025competitive}, and planning \cite{openai_research}. Central to this advancement is the emergence of reasoning and reflection capabilities within these models. These capabilities enable LLMs to iteratively refine their approach, moving closer to the correct solution by building upon their previous thought trajectories \cite{openaio3,deepseekai2025deepseekr1incentivizingreasoningcapability,gemini25pro}.

Proving complex mathematical theorems, which involves long chains of mathematical computation and logical deduction, is widely regarded as a pinnacle of human intellect \cite{deepmindalphaproof,openaiformalwebsite}. Consequently, this task has been widely adopted to benchmark the reasoning ability of LLMs \cite{zheng2021minif2f,liu2025combibench}. A common approach for LLMs tackling this task is to emulate human reasoning by generating proofs in natural language. However, verifying the correctness of proofs generated in this manner poses a significant challenge -- identifying logical fallacies or subtle errors in natural language mathematical arguments can be as difficult as constructing the proof itself \cite{lightman2023lets}. This verification difficulty is exacerbated by the prevalence of outcome-based reward modeling (ORM) in training state-of-the-art LLMs \cite{deepseekai2025deepseekr1incentivizingreasoningcapability,openaio3}. ORM necessitates reliable verification not just for final benchmarking, but extensively throughout the training process to generate reward signals, demanding significantly more verification effort compared to simple benchmarking runs.

Formal theorem proving systems, such as Lean \cite{moura2021lean}, Isabelle \cite{Isabelle:project}, and Coq \cite{Coq:project}, have been leveraged to address the aforementioned verification challenge. Specifically, these systems enable mathematicians to write proofs in a formal language. Corresponding proof assistants or kernels can then automatically verify the correctness of these formal proofs, often by checking if the associated code compiles or executes without error. Such an approach renders the verification process automatic and aligns well with the outcome-based reward modeling methods mentioned above, offering a scalable solution to the verification bottleneck.

However, validating proofs generated by LLMs using formal environments also presents challenges. The primary difficulty lies in effectively teaching LLMs to generate code in these formal languages. As languages specifically designed for proof verification, formal languages often exhibit syntactic structures and paradigms that are significantly different from common programming languages like C++ or Python, making them effectively `out-of-domain' for models primarily trained on general code. Furthermore, the distribution of formal language data within large code corpora is typically sparse. For example, LEAN-GitHub \cite{wu2024lean}, currently the largest Lean 4 codebase, contains only 0.13B tokens, representing a minuscule fraction 0.1\% of the typical code corpora \cite{huang2024opencoder,zhu2024deepseek}, which contain around hundreds of billions of tokens. Consequently, general-purpose LLMs, such as DeepSeek R1 \cite{deepseekai2025deepseekr1incentivizingreasoningcapability}, Gemini 2.5 Pro \cite{gemini25pro}, and OpenAI o3 \cite{openaio3}, often exhibit limited proficiency in formal languages due to this data scarcity \cite{kimina_prover_2025}.

Moreover, developing specialized models via fine-tuning presents its own hurdles. Acquiring high-quality labeled data for formal theorem proving is expensive and challenging, as it requires annotators with rare expertise in both advanced mathematics and the specific formal language, and the volume of data needed can be substantial. This is compounded by the significant computational cost associated with training or fine-tuning large-scale models \cite{kimina_prover_2025,RenEtAl2025:DeepSeekProverV2}.

\textbf{Our Contribution.} In contrast to approaches demanding extensive data labeling or task-specific fine-tuning, we explore a novel framework that leverages general-purpose LLMs~\citep{deepseekai2025deepseekr1incentivizingreasoningcapability,openaio3,gemini25pro} and a formal proof assistant environment using Lean 4. We propose an agent-based framework, Delta Prover\footnote{"Delta" stands for "\textbf{De}composition and iterative ref\textbf{l}ec\textbf{t}ion \textbf{a}gent"}, designed to harness the inherent reasoning and reflection capabilities of LLMs through effective orchestration, thereby eliminating the need for bespoke training data or model adaptation.

Delta Prover employs an LLM-in-the-loop agent that actively participates in the theorem-proving process. By iteratively interacting with the Lean 4 environment, this agent transforms the LLM from a passive generator into a dynamic partner in formal reasoning.
The agent has two primary components: a novel algorithmic framework and a specialized Lean 4 syntax. The syntax provides the scaffolding for the entire process, managing task decomposition, storing the state of sub-problems, and assembling solved steps into a final, verified proof. The framework, in turn, is built upon two core mechanisms:
(a) \textbf{Reflective Decomposition}: This leverages the LLM's capacity for high-level strategy to break down complex theorems into simpler sub-problems. Crucially, the agent can reflect on failed decompositions to revise its approach.
(b) \textbf{Iterative Proof Repair}: This creates a tight feedback loop where errors from the Lean 4 verifier directly guide the LLM’s next attempt, a process augmented by automated theorem retrieval to find relevant lemmas.
Together, these elements enable the agent to systematically tackle complex proofs, learn from its mistakes, and produce fully machine-verifiable results in Lean 4.

Building on the techniques previously described, \textbf{we empirically prove that general-purpose LLMs, when employed within a prover agent, can achieve remarkable theorem-proving prowess}, effectively challenging the established dominance of bespoke models in the field.
Specifically:
\begin{itemize}
    \item As demonstrated in Figure~\ref{fig:comp_result}, Delta Prover achieves a 95.9\% success rate on the miniF2F-test benchmark, establishing a new state-of-the-art that surpasses existing provers, including those specifically fine-tuned. Notably, on the more challenging IMO problems within miniF2F-test, Delta Prover attains an 85\% success rate, exceeding the previous state-of-the-art~\cite{RenEtAl2025:DeepSeekProverV2}.
    \item Our ablation studies reveal that Delta Prover possesses significantly stronger test-time scaling compared to the widely applied best-of-N sampling strategy.
\end{itemize}
This agent-centric methodology not only unlocks more potent and adaptable automated reasoning capabilities but also markedly curtails the effort involved in data collection and fine-tuning bespoke models. Furthermore, our findings offer valuable insights into harnessing LLM reflection and reasoning to construct highly capable agents.

\begin{figure}[htbp]
    \centering
    \includegraphics[width=1.0\textwidth]{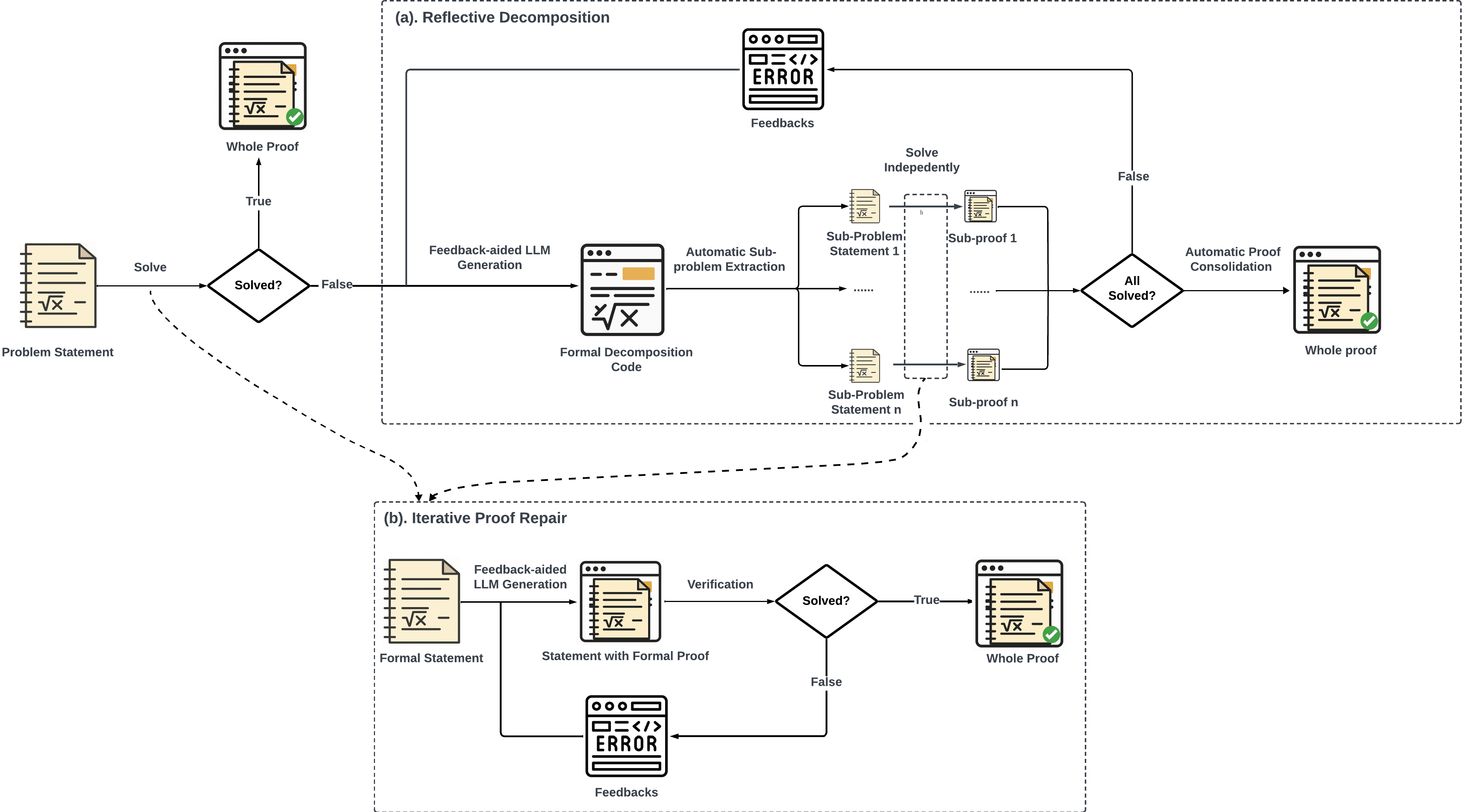}  % 替换为你的图片文件名
    \caption{An overview of our agent-driven prover framework. The system features the interplay between (a). a Reflective Decomposition process and (b). an iterative proof repair strategy. }
    \label{fig:main}
\end{figure}

\section{Approach}

% Our agent-driven Lean4 prover operates as an iterative loop, seamlessly integrating a general-purpose LLM with the Lean4 environment. As illustrated in Figure \ref{fig:main}, our prover has three key pillars:

% \begin{enumerate}
%     \item \textbf{Hierarchical Decomposition via Extended Syntax}: a custom DSL (domain specific language) lets the LLM introduce and manage subgoals and proof placeholders.  Once all subgoals are solved, the DSL will automatically integrate the proof of all subgoals in the whole proof of the original goal.
    
%     \item \textbf{Proof Sketch Generation}: the LLM produces and self‑evaluates high‑level proof outlines. To fully leverage the capability of LLM in reasoning using natural language, we first let LLM to generate and select NL proof, and then let LLM write formal decomposition code based on the NL proof. 
    
%     \item \textbf{Proof Generation via iterative error correction}: In our proof-generation loop, unlike traditional BoN approaches, we iteratively guide the LLM to refine its proof based on feedback from Lean4 until the proof is error-free. Specifically, we provide three forms of detailed feedback to the model: (1) the current tactic state and compiler error messages, (2) documentation strings for relevant tactics, and (3) the code and type information of pertinent theorems from mathlib4.
% \end{enumerate}

General-purpose Large Language Models (LLMs) demonstrate impressive capabilities when faced with natural language mathematical problems. However, their direct application to formal theorem proving faces two significant drawbacks that limit their effectiveness: 
first, formal proofs demand complete correctness, a stringent requirement often unmet as the probability of LLM-generated errors tends to grow exponentially with the proof length; 
second, current approaches often deploy LLMs to sample proof steps or entire proofs in parallel \cite{kimina_prover_2025,RenEtAl2025:DeepSeekProverV2}, but the (conditionally) independent nature of these samples limits effective scaling and efficient discovery of a complete, valid proof.

% To address these issues, we introduce an agent-driven Lean 4 prover operating as an iterative loop, seamlessly integrating a general-purpose LLM with the Lean 4 environment. As illustrated in Figure~\ref{fig:main}, our prover is built upon two key pillars designed to counter the aforementioned challenges:

% \begin{enumerate}
%     \item \textbf{Proof Sketch Generation:} The LLM produces and self-evaluates high-level proof outlines. To fully leverage the LLM's natural language reasoning strengths, we first make the LLM to generate and select promising proof sketches in natural language (NL). Subsequently, the chosen NL sketch is translated by the LLM into a formal decomposition structure.

%     \item \textbf{Proof Generation via Iterative Error Correction:} The LLM iteratively refines LLM outputs using fix-up information until the proof is complete and error-free. If the LLM fails to solve the problem in its first attempt, we will further aid it with the feedback from the Lean 4 environment and retrieved possible tactic and theorem candidates. 
% \end{enumerate}

To address these issues, we introduce an agent-driven Lean 4 prover, Reflection Agent Prover (Delta Prover), which operates in an iterative loop and seamlessly integrates a general-purpose LLM with the Lean 4 environment. As illustrated in Figure~\ref{fig:main}, our prover builds upon the emerged reflection and reasoning ability of recent general-purpose LLMs, and relies on two key pillars designed to overcome the aforementioned challenges:
\begin{itemize}
    \item \textbf{Iterative Proof Repair:} The LLM repeatedly attempts formal proof steps, progressively refining its output based on textual feedback from the Lean 4 environment until a complete and verified proof is achieved. Should an attempt fail, Delta Prover provides the LLM with diagnostic information from Lean 4, alongside retrieved relevant tactics and theorems, to guide subsequent iterations.
    \item \textbf{Reflective Decomposition:} For complex problems, the LLM initially formulates a high-level proof sketch using a novel Domain-Specific Language (DSL, detailed below), guided by a natural language outline. Delta Prover then systematically decomposes this formal sketch into sub-problems via the DSL. Each sub-problem is addressed using the iterative proof repair mechanism. Should a sub-problem prove intractable, the LLM reflects on the decomposition, revising it or devising alternative proof strategies.
\end{itemize}

To support this framework, we leverage a \textbf{Decomposition via Extended Syntax} methodology. 
This involves a custom Domain-Specific Language (DSL) built atop Lean 4, enabling the LLM to create and manage subgoals annotated with proof placeholders. Once all subgoals are successfully proven, the DSL automatically integrates these intermediate proofs to construct the complete formal proof for the original goal.

The remainder of this section is organized as follows: First, we elaborate on Delta Prover's two main components: the Iterative Proof Repair loop and the Reflective Decomposition process, and describe how these approaches are deployed. We then detail the extended DSL facilitating goal management within Lean 4.

\subsection{Algorithmic Framework}
\label{sec:framework}

This section details the workings of Delta Prover. To facilitate clear descriptions in the subsequent text, we first establish the following notation. Let 
$S$
 denote a formal statement and  $s$
 its corresponding natural language version. A formal proof of 
$S$
 is denoted by 
$P$. For the decomposition stage, $p$
 represents an informal proof (or proof plan), and $D$
 signifies a formal proof sketch expressed in our domain-specific language (DSL, introduced in Section \ref{sec: playm}).
 
 % When a problem $S$
 % is decomposed, its subproblems are denoted as 
 % $S_1, S_2, \dots, S_n$. Finally, within the iterative error correction process for a given problem (or subproblem), $P^j$
 % denotes the formal  attempt at iteration $j$.

\subsubsection{Iterative Proof Repair}
\label{sec:iterative_error_correction}

Current approaches~\cite{kimina_prover_2025, xin2024deepseek, RenEtAl2025:DeepSeekProverV2}
typically parallelize Large Language Model (LLM) rollouts,
hoping that with a sufficient number of samples, independent exploration
will eventually discover a valid solution.
This strategy, however, often leads to isolated trajectories,
fails to fully leverage the capacity for reflection and reasoning, and results in an unsatisfactory test-time scaling law as
demonstrated by recent reasoning-focused models~\cite{gemini25pro,openaio3,deepseekai2025deepseekr1incentivizingreasoningcapability}.
Recognizing this limitation, our framework incorporates an iterative
error correction process as a core component, with the corresponding pseudocode provided in Algorithm \ref{alg:proof_refinement}.
Specifically, given a formal statement $S$ and its informal
counterpart $s$, this process proceeds as follows:

\textbf{Initial Solution Generation.}
The LLM is prompted with the formal statement $S$ to generate an initial formal proof candidate, $P^0$. Recognizing that general-purpose LLMs may lack proficiency in Lean 4 syntax and tactics, we  provide guidelines demonstrating:
\begin{itemize}
    \item \textbf{Formatting Conventions:} LLMs often demonstrate inconsistent application of Lean 4's code formatting rules, such as the correct syntax following various tactics (e.g., \texttt{rcases}, \texttt{cases
    }). We provide explicit instructions on proper formatting.
    \item \textbf{Effective Tactics:} We provide the doc-string of powerful Mathlib tactics like \texttt{linarith}, \texttt{ring}, and \texttt{omega}. These tactics automate significant reasoning steps within Lean 4, and providing explicit examples helps the LLM understand their appropriate application scope.
    \item \textbf{Lean 4 Specification:} LLMs frequently generate Lean 3 code when prompted for Lean proofs if not explicitly instructed otherwise. We conjecture that this is due to the relatively recent community transition from Lean 3 to Lean 4, and thus many Lean 3 codes remain in the knowledge base of LLMs. This Lean 3 code is incompatible with Lean 4 verification. Therefore, we emphasize in the prompt that the generated proof must adhere strictly to Lean 4 syntax and libraries.
\end{itemize}
These guidelines aim to improve the quality and correctness of the initial proof $P^0$. If the statement-proof pair $(S, P^0)$ successfully validates against the Lean 4 kernel, the process terminates successfully for this problem. Otherwise, we proceed to the correction phase.

\textbf{Feedback Augmented Proof Repair.}
If a proof attempt $P^i$ fails validation, the Lean 4 kernel provides error messages indicating the location and type of the first error, sometimes suggesting potential fixes. While useful, this message alone often lacks sufficient context for effective correction. Therefore, we augment the prompt for the next iteration ($i+1$) with additional information:
\begin{itemize}
    \item \textbf{Failed Proof:} The incorrect proof $P^i$.
    \item \textbf{Message from Lean 4 kernel:} The tactic state and Lean 4 Error Message from Lean 4 kernel.
    \item \textbf{Retrieved Information:} We observe that LLMs may know the informal step but struggle to find the exact name and signature of a formal theorem or tactic from Mathlib4, which is a large and complex library. If the error involves an unknown or incorrect identifier, we retrieve relevant theorems/definitions based on the errored name and include them in the prompt.
\end{itemize}
This comprehensive prompt (please refer to the concrete template in Figure \ref{fig:prompt solve}) is fed to the LLM to generate a revised proof, $P^{i+1}$. If $(S, P^{i+1})$ validates, the process succeeds. Otherwise, the fixup cycle repeats. If a valid proof is not found after a predetermined maximum number of iterations, this direct approach is deemed unsuccessful.

% \begin{algorithm}
% \caption{Iterative Proof Refinement Algorithm}
% \label{alg:proof_refinement}
% \begin{algorithmic}[1] % The [1] enables line numbering

% \Require Formal statement $S$, informal statement $s$, LLM $\Phi$, retrieval model $\Psi$, number of rounds $m$, number of iterative refinements per round $n$, Lean 4 kernel $\mathcal{L}$, instruction for formal proof $I_{fp}$
% \Ensure A formal proof $P$ or \texttt{Failure}

% \For{$i \gets 0 \text{ to } m-1$} \Comment{Loop for $m$ rounds}
%     \State $P \gets \Phi(S, s, I_{fp})$ \Comment{Generate initial proof candidate}
%     \For{$j \gets 0 \text{ to } n-1$} \Comment{Loop for $n$ iterative refinements}
%         \State $E \gets \mathcal{L}(S, P)$ \Comment{Verify $P$ using Lean 4 kernel}
%         \If{$E \text{ contains no error}$}
%             \State \Return $P$ \Comment{Successfully verified proof found}
%         \Else
%             \State $T \gets \Psi(S, s, P, E)$ \Comment{Retrieve theorems/definitions $T$}
%             \State $P \gets \Phi(S, s, I_{fp}, E, T)$ \Comment{Refine proof based on error $E$ and retrieved information $T$}
%         \EndIf
%     \EndFor
% \EndFor
% \State \Return \texttt{Failure} \Comment{No verifiable proof found after $m$ rounds and $n$ refinements per round}

% \end{algorithmic}
% \end{algorithm}

\begin{algorithm}
\caption{\texttt{Iterative Proof Repair}}
\label{alg:proof_refinement}
\begin{algorithmic}[1] % The [1] enables line numbering

\Require Formal statement $S$; Informal statement $s$; LLM $\Phi$; Retrieval model $\Psi$; Number of rounds $m$; Number of iterative repair per round $n$, Lean 4 Kernel \& DSL Environment $\mathcal{L}$
\Ensure A formal proof $P$ or \texttt{Failure}

\State Let $I_{fp}$ be the instruction for formal proof. % Moved instruction here

\If{$s$ is \texttt{None}}
    \State $s\gets \Phi(S)$ \Comment{Generate the informal statement $s$ if not provided}
\EndIf

\For{$i \gets 0 \text{ to } m-1$} \Comment{Loop for $m$ rounds}
    \State $P \gets \Phi(S, s, I_{fp})$ \Comment{Generate initial proof candidate}
    \For{$j \gets 0 \text{ to } n-1$} \Comment{Loop for $n$ iterative repairs}
        \State $\textit{kernelOutput} \gets \mathcal{L}(S, P)$ \Comment{Verify $P$ using Lean 4 kernel}
        \If{$\textit{kernelOutput} \text{ contains no error}$}
            \State \Return $P$ \Comment{Successfully verified proof found}
        \Else
            \State $\textit{retrievedInfo} \gets \Psi(S, s, P, \textit{kernelOutput})$ \Comment{Retrieve theorems/definitions}
            \State $P \gets \Phi(S, s, I_{fp}, \textit{kernelOutput}, \textit{retrievedInfo})$ \Comment{Repair proof based on error and retrieved information}
        \EndIf
    \EndFor
\EndFor
\State \Return \texttt{Failure} \Comment{No verifiable proof found after $m$ rounds and $n$ repairs per round}

\end{algorithmic}
\end{algorithm}

\subsubsection{Reflective Decomposition}
\label{sec:sketch_generation}

The iterative error correction mechanism effectively leverages the reflective capabilities of LLMs to derive correct proofs.
However, as problem difficulty escalates, the complexity of corresponding proofs also increases, potentially leading to an exponential accumulation of errors throughout the reasoning process.
Consequently, even with this error correction, LLMs may still require an excessive number of trials to converge on a correct proof.

A typical approach to resolve this challenge is Draft, Sketch, and Prove (DSP) introduced in \cite{jiang2022draft}, which breaks down the original problem into easier sub-problems, and solves sub-problems instead.
While this approach appears straightforward and standard, its implementation presents several challenges:
\begin{itemize} % leftmargin=* makes the list indent less, adjust as needed
    \item \textbf{Management of sub-problems.}  DSP uses the formal language Isabella, while current Lean 4 lack robust support for managing decomposed sub-problems. Specifically, there is no convenient way to simultaneously: (i) store decomposed sub-problems; (ii) extract them as formal statements; and (iii) integrate their individual proofs back into a unified whole proof. For instance, using the \texttt{have} tactic is inadequate for (iii)~\cite{RenEtAl2025:DeepSeekProverV2}, while listing sub-problems as lemmas falls short for (ii).
 
    \item \textbf{Proper decomposition.}  LLMs typically can not precisely estimate to which extent of difficulty the sub-problem can be solved, and if the decomposition is not proper, some decomposed sub-problems will again become the proof bottleneck. 
\end{itemize}

% To resolve the first challenge (sub-problem management), we will introduce our Domain-Specific Language (DSL) for problem decomposition in Section~\ref{sec: playm}.
% For now, it is sufficient to understand that this DSL facilitates improved sub-problem management.
% The design of our pipeline primarily addresses the second challenge (proper decomposition). This process involves: first, generating an informal proof plan and translating this plan into a formal sketch in our DSL; secondly, extract sub-problems through the DSL and call the iterative proof repair algorithm to solve each sub-problem independently; and finally, if any of the subproblems are not solved, we let the LLM to regenerate the formal sketch by providing it with which subproblems are not solved.
% These components are illustrated sequentially below, with the corresponding
% pseudocode provided in Algorithm \ref{alg:informal_guided_sketch_gen}.

To address the first challenge, sub-problem management, we introduce a Domain-Specific Language (DSL) for problem decomposition, detailed in Section~\ref{sec: playm}. For now, it suffices to note that this DSL enhances sub-problem management.
Our framework design primarily tackles the second challenge: proper decomposition. This process involves: (1) generating an informal proof plan and translating it into a formal sketch using our DSL; (2) extracting sub-problems via the DSL and attempting to solve each independently with the Iterative Proof Repair loop; and (3) if any sub-problems remain unsolved, prompting the LLM to regenerate the formal sketch, informed by the list of unsolved sub-problems.
These steps are illustrated below, with corresponding pseudocode in Algorithm~\ref{alg:informal_guided_sketch_gen}.

\textbf{Initial Formal Sketch Construction.} We first construct the informal proof plan $p$ by prompting the LLM with the statements $(S,s)$ and specific guidance to generate $p$ in a step-by-step format, conducive to later translation into a formal sketch (please refer to Figure \ref{fig:prompt nl} for the prompt template).

Given the statements $(S, s)$ and the informal proof plan $p$, the LLM is then tasked with generating a corresponding formal proof sketch $D$ in our custom Domain-Specific Language (DSL) (detailed in Section~\ref{sec: playm}).
Our DSL is intentionally designed so that the structure of the formal proof sketch $D$ closely mirrors that of the informal plan $p$. Consequently, this generation process is analogous to the auto-formalization task~\cite{wu2022autoformalization}.
To guide the LLM towards accurate formalization, we incorporate the following elements into the prompt (see a concrete template in Figure~\ref{fig:prompt decomp}):
\begin{itemize}
    \item \textbf{DSL Format Guidelines:} Since apparently our DSL is not part of standard LLM pre-training data, directly prompting for code generation in our DSL would likely lead to errors. We mitigate this by providing in-context examples that illustrate the syntax and usage of each DSL construct, with particular emphasis on the custom tactics introduced in Section~\ref{sec: playm}.
    \item \textbf{Auto-formalization Pitfalls:} The translation from informal mathematical language to formal statements can harbor subtle traps, such as implicit assumptions or incorrect quantifier scope. To address these, the prompt explicitly highlights common pitfalls and includes examples of correct formalization.
\end{itemize}
The final output of this step is the formal proof sketch $D$.

% \textbf{Sub-problem Extraction and Solve.} As detailed in Section~\ref{sec: playm} and "Subproblem extraction and proving" part in Figure \ref{fig:playm}, our DSL (specifically, the \texttt{show ... by} tactic) can parse $D$ and automatically extract the formal statements for each required sub-proof, either yielding the subproblems $S_1, \dots, S_n$ or an illegal formal draft. Each sub-problem is then sent to the Iterative Proof Repairment pipeline described in the last section.

% \textbf{Iterative Decomposition Refinement.} If any of the decomposed sub-problems failed to be solved, we record the unsolved sub-problems and combine this information to let the LLM regenerate a proof sketch $D$ with a more proper grid. The pipeline then proceeds until a certain threshold of decomposition attempts is reached or all sub-problems are solved.

\textbf{Sub-problem Extraction and Solving.} Following the methodology detailed in Section~\ref{sec: playm} and illustrated in the "Subproblem extraction and proving" segment of Figure~\ref{fig:playm}, our Domain-Specific Language (DSL) employs its \texttt{show ... by} tactic to parse the initial proof draft $D$. This tactic automatically extracts formal statements for all required sub-proofs. The outcome is either a collection of sub-problems, $S_1, \dots, S_n$, or the classification of $D$ as an illegal formal draft. Valid sub-problems are then passed to the Iterative Proof Repair loop.

\textbf{Iterative Decomposition Repair.} Should any sub-problems prove unsolvable by the Iterative Proof Repair loop, these failures are logged. This feedback, encompassing the list of unsolved sub-problems, directs the LLM to regenerate the proof sketch $D$ with an improved decomposition strategy (e.g., a more appropriate "grid"). This repair loop continues until all sub-problems are solved or a maximum number of decomposition attempts is exhausted.

\begin{algorithm}
    \caption{\texttt{Reflective Decomposition}}
    \label{alg:informal_guided_sketch_gen} % Changed label to be more descriptive
\begin{algorithmic}[1] % The [1] enables line numbering

\Require
    Formal statement $S$;
    Informal statement $s$;
    LLM $\Phi$;
    Retrieval model $\Psi$;
    Lean 4 Kernel \& DSL Environment $\mathcal{L}$;
    Max decomposition attempts $l_{max}$; % Renamed for clarity
    Sub-problem solving rounds $\tilde{m}$;
    Sub-problem repair iterations $\tilde{n}$

\Ensure A list of (sub-problem, proof) pairs: $((S'_1, p'_1), \dots, (S'_k, p'_k))$, or \texttt{Failure}

\State Let $I_{ip}$ be the instruction template for informal proof generation
\State Let $I_{fs}$ be the instruction template for formal sketch generation

\State $p_{informal} \gets \Phi(S, s, I_{ip})$ \Comment{Generate initial informal proof sketch}
\State $D \gets \Phi(S, s, p_{informal}, I_{fs})$ \Comment{Generate initial formal sketch from $p_{informal}$}

\For{$i \gets 1 \text{ to } l_{max}$}
    \State $\textit{kernelOutput} \gets \mathcal{L}(D)$ \Comment{Verify formal sketch $D$; $\textit{kernelOutput}$ contains errors or sub-problems}

    \If{$\textit{kernelOutput}$ indicates a kernel-level error}
        \State $E \gets \textit{kernelOutput}$ \Comment{Feedback $E$ is the set of kernel errors}
    \Else
        \State $(S'_1, \dots, S'_k) \gets \textit{kernelOutput}$ \Comment{Extract sub-problems}
        \State $\textit{UnsolvedSubproblems} \gets [~]$
        \State $\textit{SolvedProofs} \gets [~]$
        \For{$j \gets 1 \text{ to } k$} \Comment{Attempt to solve each sub-problem}
            \State $P'_j \gets \texttt{Iterative Proof Repair}(S'_j, S, \texttt{None}, \Phi, \Psi, \mathcal{L}, \tilde{m}, \tilde{n})$
            \Comment{$S$ is overall context, \texttt{None} for initial sub-proof attempt}
            \If{$P'_j = \texttt{Failure}$}
                \State Add $S'_j$ to $\textit{UnsolvedSubproblems}$
            \Else
                \State Add $(S'_j, P'_j)$ to $\textit{SolvedProofs}$
            \EndIf
        \EndFor

        \If{$\textit{UnsolvedSubproblems}$ is empty}
            \State \Return $\textit{SolvedProofs}$ \Comment{All sub-problems solved}
        \Else
            \State $E \gets \textit{UnsolvedSubproblems}$ \Comment{Feedback $E$ is the list of unsolved sub-problems}
        \EndIf
    \EndIf

    % If we reach here, D was invalid or led to unprovable sub-problems. Refine D.
    \If{$i < l_{max}$} \Comment{Refine $D$ if not the last attempt}
        \State $D \gets \Phi(S, s, D, E)$ \Comment{Regenerate formal sketch $D$ using feedback $E$}
    \EndIf
\EndFor

\State \Return \texttt{Failure} \Comment{Maximum decomposition attempts reached or all failed}
\end{algorithmic}
\end{algorithm}

\subsubsection{The Overall Framework}

Building upon the previously detailed Iterative Proof Repair loop (Algorithm~\ref{alg:proof_refinement}) and Reflective Decomposition approach (Algorithm~\ref{alg:informal_guided_sketch_gen}), we now describe their integration into our final framework, Delta Prover.

When presented with a problem, defined by a statement pair $(S,s)$, Delta Prover initially attempts a direct solution using through Iterative Proof Repair.
If this direct approach is unsuccessful, the framework then employs Reflective Decomposition, which returns either \texttt{Failure} or a list of (sub-problem, proof) pairs. In the former case, Delta Prover terminates and returns \textbf{Failure}. In the latter, Delta Prover performs \textbf{Automatic Proof Consolidation} (illustrated in the "Proof consolidation" section of Figure~\ref{fig:playm}).
During this consolidation stage, the derived sub-proofs $P_i$ are substituted back into the DSL-based sketch.
Subsequently, necessary concluding commands (e.g., a \texttt{conclude} tactic), as dictated by the DSL's structure, are appended to produce the complete formal proof $P$ for the original statement $S$.

The pseudocode for this comprehensive framework is provided in Algorithm~\ref{alg:overall_pipeline}.

\begin{algorithm}
\caption{The Framework of Delta Prover}
\label{alg:overall_pipeline}
\begin{algorithmic}[1] % [1] enables line numbering

\Require
    Formal statement $S$;
    Informal statement $s$;
    LLM $\Phi$;
    Retrieval model $\Psi$;
    Lean 4 Kernel \& DSL Environment $\mathcal{L}$;
    Maximum decomposition attempts $l_{\text{max}}$;
    Rounds for direct proof attempt $m$;
    Iterative repairs per round (for direct proof) $n$;
    Rounds for sub-problem proofs $\tilde{m}$;
    Iterative repairs per round (for sub-problems) $\tilde{n}$ 
\Ensure A formal proof $P$ for $S$, or \texttt{Failure}

\State $P \gets \texttt{Iterative Proof Repair}(S, \texttt{None}, \Phi, \Psi, \mathcal{L}, m, n)$ \Comment{Attempt direct proof of $S$}
\If {$P \text{ is not } \texttt{Failure}$}
    \State \Return $P$
\EndIf

\Comment{Direct proof failed; attempt Reflective Decomposition}
\State $\textit{decompResult} \gets \texttt{Reflective Decomposition}(S, s, \Phi, \Psi, \mathcal{L}, l_{\text{max}}, \tilde{m}, \tilde{n})$
\Comment{\textit{decompResult} is a list of (sub-problem, proof) pairs or \texttt{Failure}}

\If {$\textit{decompResult} \text{ is not } \texttt{Failure}$}
    \Comment{Automatic Proof Consolidation Stage}
    \State $P_{\text{consolidated}} \gets \mathcal{L}(\textit{decompResult})$
    \Comment{Consolidate sub-proofs with the DSL sketch via $\mathcal{L}$ }
    \If {$P_{\text{consolidated}} \text{ is not } \texttt{Failure}$}
        \State \Return $P_{\text{consolidated}}$
    \EndIf
\EndIf

\State \Return \texttt{Failure} \Comment{All approaches failed}

% --- Start of commented out section (original) ---
% \State  $D,~ (S_1, \dots, S_n)\gets \text{Algorithm \ref{alg:informal_guided_sketch_gen}}~(S, s, \dots)$
% \Comment{Attempt to decompose the problem}

%     \If{$n=0$}
%         \State \Return \texttt{Failure} \Comment{Decomposition yielded no subproblems (sketch $D$ did not decompose $S$)}
%     \Else
%         \State $P_{sub\_proofs} \gets [~]$ \Comment{Initialize list for sub-proofs}
%         \State all\_subproblems\_solved $\gets \texttt{true}$
%         \For{$i \gets 1 \text{ to } n$}
%             \State $P_i \gets \text{Algorithm \ref{alg:overall_pipeline}}~(S_i, \texttt{None}, d_c+1, \Phi, \mathcal{L}, d_{max})$ \Comment{Recursively solve subproblem $S_i$}
%             \If{$P_i \text{ is } \texttt{Failure}$}
%                 \State all\_subproblems\_solved $\gets \texttt{false}$
%                 \State \textbf{break} \Comment{Stop if any subproblem fails}
%             \EndIf
%             \State Add $P_i$ to $P_{sub\_proofs}$
%         \EndFor

%         \If{all\_subproblems\_solved}
%             \State $P_{consolidated} \gets \mathcal{L}(D, P_{sub\_proofs})$ \Comment{Consolidate sub-proofs using sketch $D$ and kernel/DSL}

%                  \State \Return $P_{consolidated}$
%         \Else
%             \State \Return \texttt{Failure} \Comment{Failed to solve one or more subproblems}
%         \EndIf
%     \EndIf
% --- End of commented out section ---

\end{algorithmic}
\end{algorithm}

\subsection{Decomposition via Extended Syntax}
\label{sec: playm}
The previous section introduced our use of a Domain-Specific Language (DSL) to address challenges in managing sub-problems within Lean 4.
We noted that while native Lean 4 syntax is powerful for low-level tactics, it is less suited for high-level proof sketching, isolating subproblem contexts, and automatically reintegrating subproofs.
This section delves into the details of our DSL.

Our DSL is implemented by leveraging Lean 4's metaprogramming facilities.
Specifically, we introduce an additional monad layer, \texttt{PlayM}, built upon \texttt{TacticM}, to record and manage intermediate proof states.
These saved states are subsequently converted into complete and valid Lean 4\ proof scripts using Lean 4’s built-in delaborator.

\begin{figure}[htbp]
    \centering
    \includegraphics[width=1.0\textwidth]{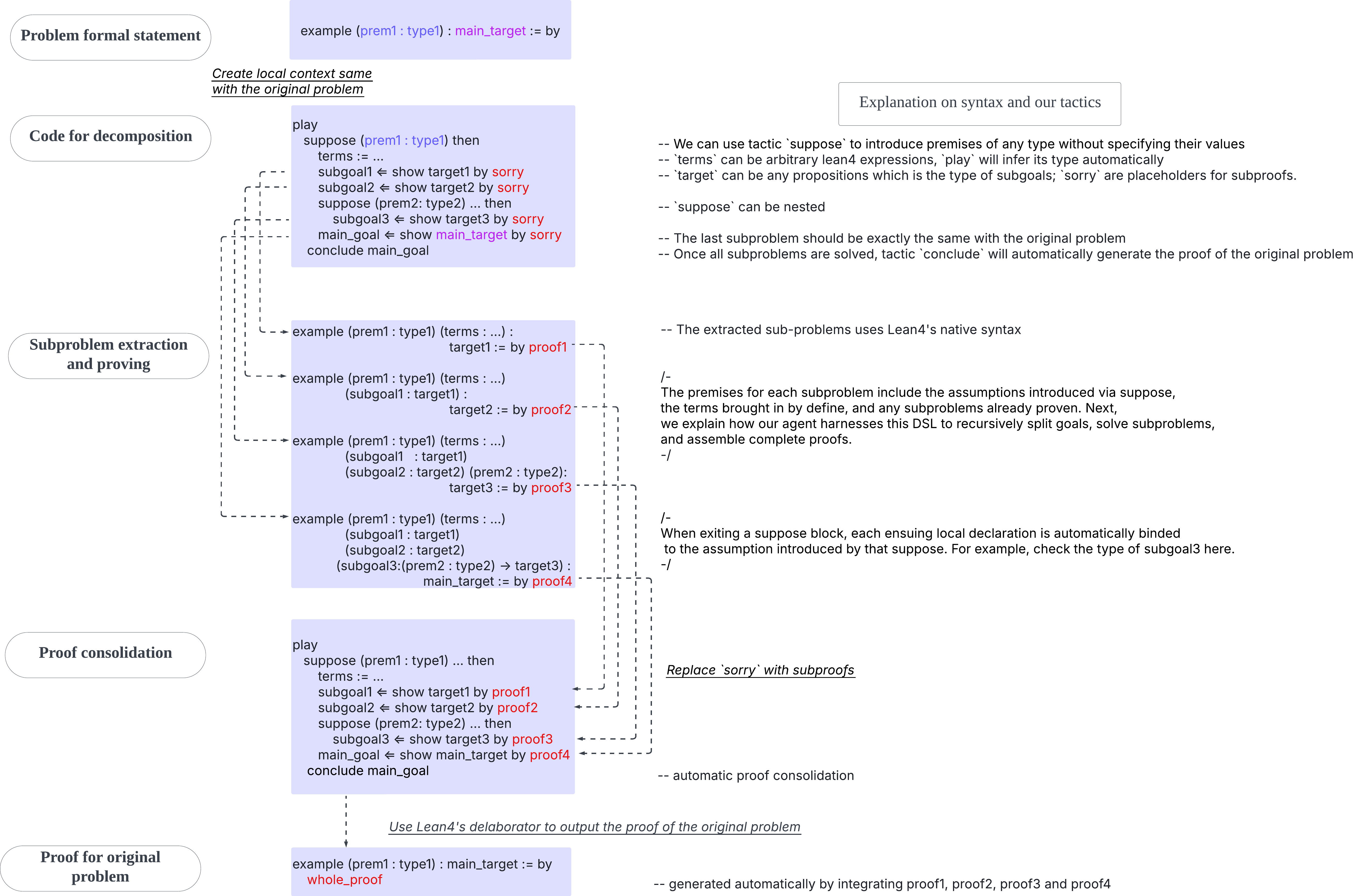}  % 替换为你的图片文件名
    \caption{A demonstration of our Domain Specific Language (DSL), built upon Lean 4, and its role in managing decomposed sub-problems. Specifically, the DSL facilitates: maintaining the current state of sub-problem decomposition, extracting sub-problems into formal statements, and consolidating the proofs of sub-problems into the proof of the original problem.}
    \label{fig:playm}
\end{figure}

To facilitate more effective decomposition, we have implemented four specialized tactics in PlayM. Figure \ref{fig:playm} illustrates syntaxes of these tactics and how these tactics are employed to systematically split a theorem into subproblems and subsequently reassemble the solutions into a coherent proof. 
In a nutshell, our DSL provides four key capabilities:

\begin{enumerate}
    \item Tactic \textbf{Suppose} for Hypothesis Introduction: we use this tactic to  introduce new hypotheses into the proof context. It allows the assumption of arbitrary Lean 4 types as premises without the need to explicitly specify their values.
    \item Tactic \textbf{Define} for Arbitrary Expression Construction: The define tactic enables the introduction of arbitrary Lean 4 expressions, with playm automatically inferring their types. For example, one can use define to declare a new function from integer to integer or a Finset over natural numbers. 
    \item Tactic \textbf{ShowBy} for Subgoal Creation: The ShowBy tactic provides an interface between PlayM and TacticM for leveraging well-developed tactics from Mathlib. It allows us to explicitly pose subproblems. It functions similarly to the native Lean 4 showby tactic, with the key difference being that our implementation records the proofs generated within it. These recorded proofs are then utilized later for assembling a complete, integrated proof. 
    \item Tactic \textbf{Conclude} for Proof Consolidation: The conclude tactic is responsible for proof consolidation: within PlayM, every subproblem’s proof steps and dependency graph are recorded, and conclude tactic leverages these saved proofs—together with Lean 4’s delaborator to emit a fully  coherent Lean 4 proof for the original problem.
\end{enumerate}

Our extended DSL streamlines the workflow of Delta Prover. However, its specialized nature means general-purpose LLMs may lack familiarity and struggle to generate appropriate code. To overcome this, we meticulously designed prompts to guide the LLM in using the DSL effectively for both problem decomposition and solving, as elaborated in Section \ref{sec:sketch_generation}.

\section{Related Work}
\textbf{Specialized LLMs for Formal Theorem Proving.}
Developing specialized models for formal theorem proving by training on large, proof-focused datasets has been a major research direction. Early efforts \citep{whalen2016holophrasm,irving2016deepmath,wang2017premise,selsam2018learning,loos2017deep,bansal2019learning,rabe2020mathematical,huang2018gamepad,yang2019learning,wang2020learning} explored diverse tasks and formalisms, but were often hampered by limitations in model scale, architecture, and data availability, restricting their performance or scope, sometimes only addressing sub-tasks like premise selection.

A turning point came with GPT-f \citep{polu2020generative}, which demonstrated the power of generative pretraining and model scaling. Subsequent work built upon this, introducing key techniques and resources: the challenging miniF2F benchmark \citep{zheng2021minif2f}, data augmentation via kernel information extraction \citep{han2021proof}, and expert iteration for continuous improvement \citep{poluformal}. The integration of tree search methods \citep{lample2022hypertree} provided another performance boost. These combined strategies significantly improved results, reaching milestones like $40.6\%$ pass@1 accuracy on miniF2F-test.
Building on these foundations (large-scale pretraining, expert iteration, tree search) and leveraging progress in base models, numerous recent provers have emerged \citep{jiang2021lisa,jiang2022thor,wu2022autoformalization,wang2023dt,first2023baldur,xin2024deepseek,wu2024internlm2,li2024hunyuanprover,lin2025goedel,xin2025bfs,zhang2025leanabell}.  Additional enhancements include integrating dedicated premise retrieval and reranking models \citep{yang2023leandojo,mikula2023magnushammer} and deploying problem generation models to enable self-play learning loops \citep{dong2025stp}. Furthermore, inspired by demonstrations of enhanced reasoning via explicit thinking steps \citep{jaech2024openai}, some systems now incorporate natural language reasoning alongside formal proof generation \citep{lin2024lean, kimina_prover_2025,RenEtAl2025:DeepSeekProverV2,kimina_prover_2025_new}.

Overall, the development of specialized, fine-tuned models represents the current state-of-the-art in automated formal theorem proving. However, achieving these results typically demands significant resources for data curation and training. This work diverges from that framework by employing general-purpose LLMs as prover agents, without specialized training. We now review prior work adopting this alternative approach.

\textbf{General-Purpose LLMs for Formal Theorem Proving.} 
Recently, the rapidly advancing capabilities of general-purpose Large Language Models (LLMs) have inspired a complementary line of research: using these powerful models in agentic roles to perform complex tasks without specialized fine-tuning. In the realm of theorem proving, DSP (Draft, Sketch, and Prove) \citep{jiang2022draft} stands out as a prominent early example. DSP treats a general-purpose LLM as a black-box reasoning engine, prompting it to first decompose an informal proof into subproblems, and then formalize and solve these subproblems within a proof assistant like Isabelle. This approach yielded strong performance without requiring any model fine-tuning. Building on this, subsequent work introduced further refinements: COPRA \citep{thakur2023context} and Lyra \citep{zheng2023lyra} iteratively augment the prompt using execution feedback; \cite{zhao2023decomposing, zhao2024subgoalxl} improved the quality of the initial informal proof decomposition to better guide formal sketch generation; LEGO-Prover \citep{wang2023lego} proposed combining DSP with a growing library of solved subproblems; and \cite{wang2024proving} applied the DSP strategy recursively, expanding the proof as a tree structure.

% Our work belongs to and is inspired by this line of work. Specifically, the iterative proof repairment pipeline is similar to CORPA in spirit, with the difference that we utilize the LLM's ability to generate whole proof while CORPA executes in a step-by-step manner. The reflective decomposition process builds upon DSP, while we novelly utilize the reflection of LLM to iteratively derive better decomposition.

Our work aligns with and draws inspiration from this established line of research. The iterative proof repair loop, for instance, is conceptually akin to CORPA. We advance this by leveraging an LLM's capacity for generating entire proofs, contrasting with CORPA's incremental execution. Similarly, our reflective decomposition process extends DSP by uniquely employing the LLM's reflective capabilities to iteratively derive more effective decomposition strategies.

% Recently, due to the ever-growing intelligence of general-purpose llms, a complementary line of works has emerged: using powerful LLMs in agentic roles to perform complex tasks without additional training. In the realm of theorem proving, DSP (Draft, Sketch, and Prove) \cite{jiang2022draft} is a prominent example. DSP treats general-purpose llms as a black-box reasoning engine, and let llms to first decompose the informal proof into subproblems, and then formalize and solve these subproblems in Isabelle. This yield state-of-the-art performance without any fine-tuning. Following this work, COPRA \cite{thakur2023context} and Lyra \cite{zheng2023lyra} which iteratively augments the prompt according to execution feedback and retrieved lemmas. \cite{zhao2023decomposing, zhao2024subgoalxl} advance DSP by refining the quality of informal proof to better inform the generation of formal proof sketches. LEGO-Prover \cite{wang2023lego} propose to combine DSP with an growing library of proved subproblems, and retrieve useful subproblems when proving new theorems. \cite{wang2024proving} proposes to apply DSP recursively, expanding the proof as a tree structure.

% Our work builds upon this paradigm, extending it with a specialized interface for Lean4 and a structured approach to handle subgoals and errors. In particular, we aim to demonstrate that a general-purpose LLM can match or exceed specialized theorem solvers by equipping it with an intelligent agent framework, bridging the gap between informal reasoning and formal proof verification.

\section{Experiments}
In this section, we provide a thorough evaluation of our prover agent. We will start with introduction of basic experiment settings, and then assess the performance of our prover agent over popular benchmarks. We end this section by demonstrating the effect of the key components of our prover through ablation studies.

\subsection{Basic Experiment Setup}

\textbf{General-Purpose LLM.} We employ Gemini 2.5 Pro 05-06 \cite{gemini25pro} as the general-purpose LLM in our agent. We choose the sampling parameters as $\texttt{temperature}=1$, agreeing with common usage.  

\textbf{Benchmarks.} We use the following benchmarks for evaluation:
\begin{itemize}
    \item \textbf{MiniF2F-test.} MiniF2F-test is the test split of the MiniF2F benchmark \cite{zheng2021minif2f}. This dataset contains 244 problems collected from mathematics competitions and the MATH dataset \cite{hendrycksmath2021} with various difficulties. We use its Lean 4 version \cite{Yang_miniF2F-lean4_2022_misc} and fix errors in the statements similar to \cite{kimina_prover_2025,RenEtAl2025:DeepSeekProverV2}.  We also consider miniF2F-test-IMO, which consists of all IMO problems within miniF2F-test, and thus works as a more challenging benchmark.
\end{itemize}

% \textbf{Lean 4.} We use Lean 4 with version xxx.

%\subsection{Hello World}
\subsection{Benchmark Performance}
% \textbf{Performance over MiniF2F-test.} Table \ref{tab:comparison} compares the performance of our prover agent over MiniF2F-test against those of state-of-art provers. Our prover agent demonstrating a new highest accuracy over this benchmark, beating strong baselines such as Kimina-Prover-Preview \cite{kimina_prover_2025} and Deepseek Prover V2 \cite{RenEtAl2025:DeepSeekProverV2}, which require heavy data collection and fine-tuning.

% Note that we do not include performances of existing training-free methods in Table \ref{tab:comparison}, although the best reported accuracy is 50 \% of LEGO-Prover \cite{wang2023lego}. This is because their results are not directly comparable with ours as they utilize general-purpose LLMs like GPT-4 (due to absent of stronger LLMs when conducting their experiments), which is weaker than the recent developed LLMs such as DeepSeek-V3. 

Table~\ref{tab:comparison} presents a performance comparison of our prover agent against state-of-the-art provers on the MiniF2F-test benchmark.
Our agent establishes a new highest accuracy on this benchmark, outperforming strong baselines including Kimina-Prover 72B~\cite{kimina_prover_2025,kimina_prover_2025_new} and DeepSeek-Prover-V2 671B~\cite{RenEtAl2025:DeepSeekProverV2}.
It is noteworthy that these competing methods require substantial data collection and fine-tuning efforts.

% Due to the serialized nature of ingredients, such as proof repairment in our approach, the sample budget in Table \ref{tab:comparison} is calculated in the way that once RAP solves a problem, we record the corresponding the concrete API call and the final sample budget is the average of all problems. Utilizing these statistics, we are also able to illustrate the test-time scaling law of RAP, as plotted in Figure \ref{fig:scaling law}.

\begin{figure}[htbp]
    \centering % 整体居中，使得两个子图作为一个整体在页面上居中

    % --- 第一张子图 ---
    \begin{subfigure}[b]{0.48\textwidth} % 子图宽度，例如文本宽度的48%
        \centering % 子图内部的图片居中
        \includegraphics[width=\textwidth]{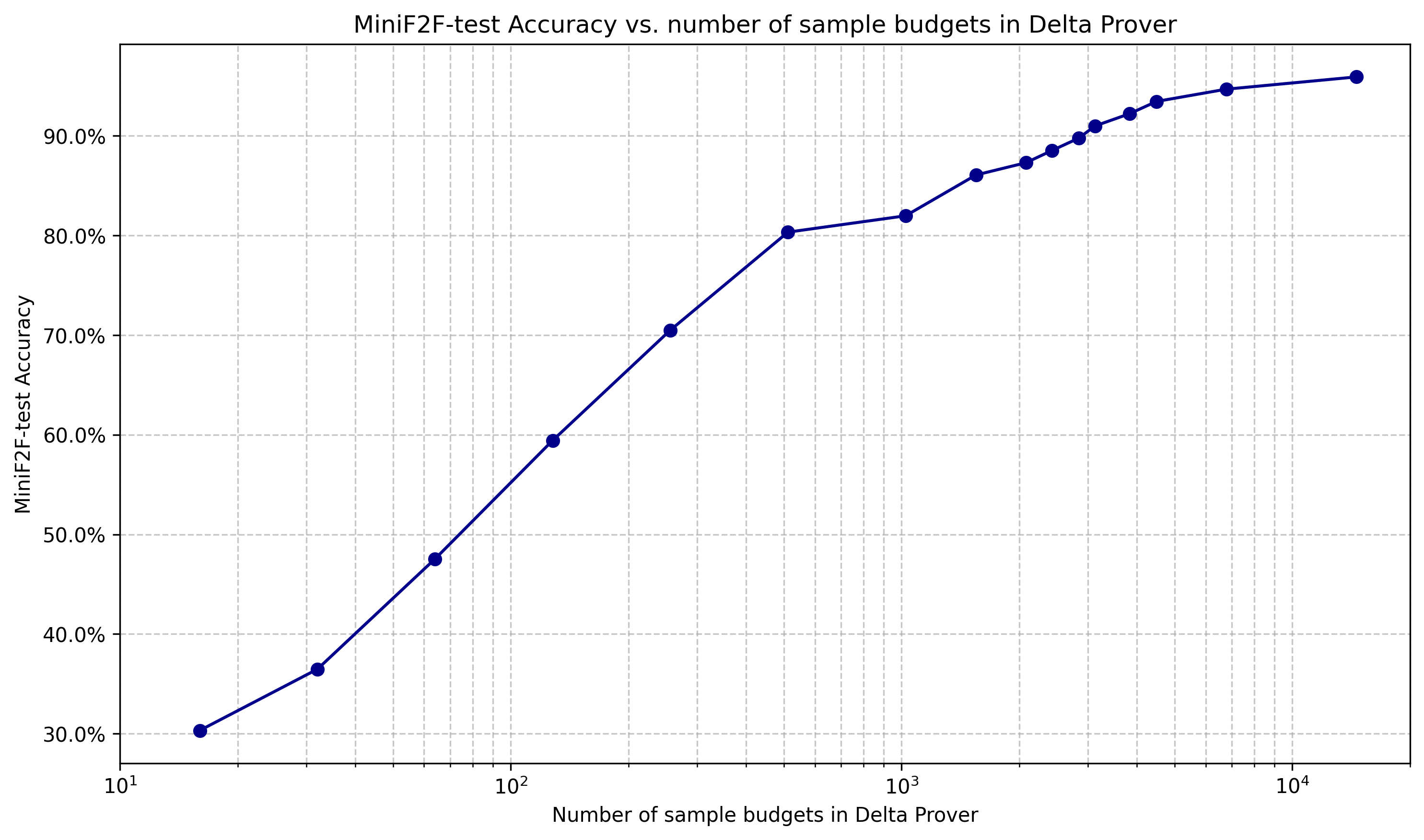} % 替换为你的第一张图片文件名
    \end{subfigure}%  <--- 注意：这里最好有一个 % 来避免潜在的意外空格导
    \hspace{3pt}
    % --- 第二张子图 ---
    \begin{subfigure}[b]{0.48\textwidth} % 子图宽度，例如文本宽度的48%
        \centering % 子图内部的图片居中
        \includegraphics[width=\textwidth]{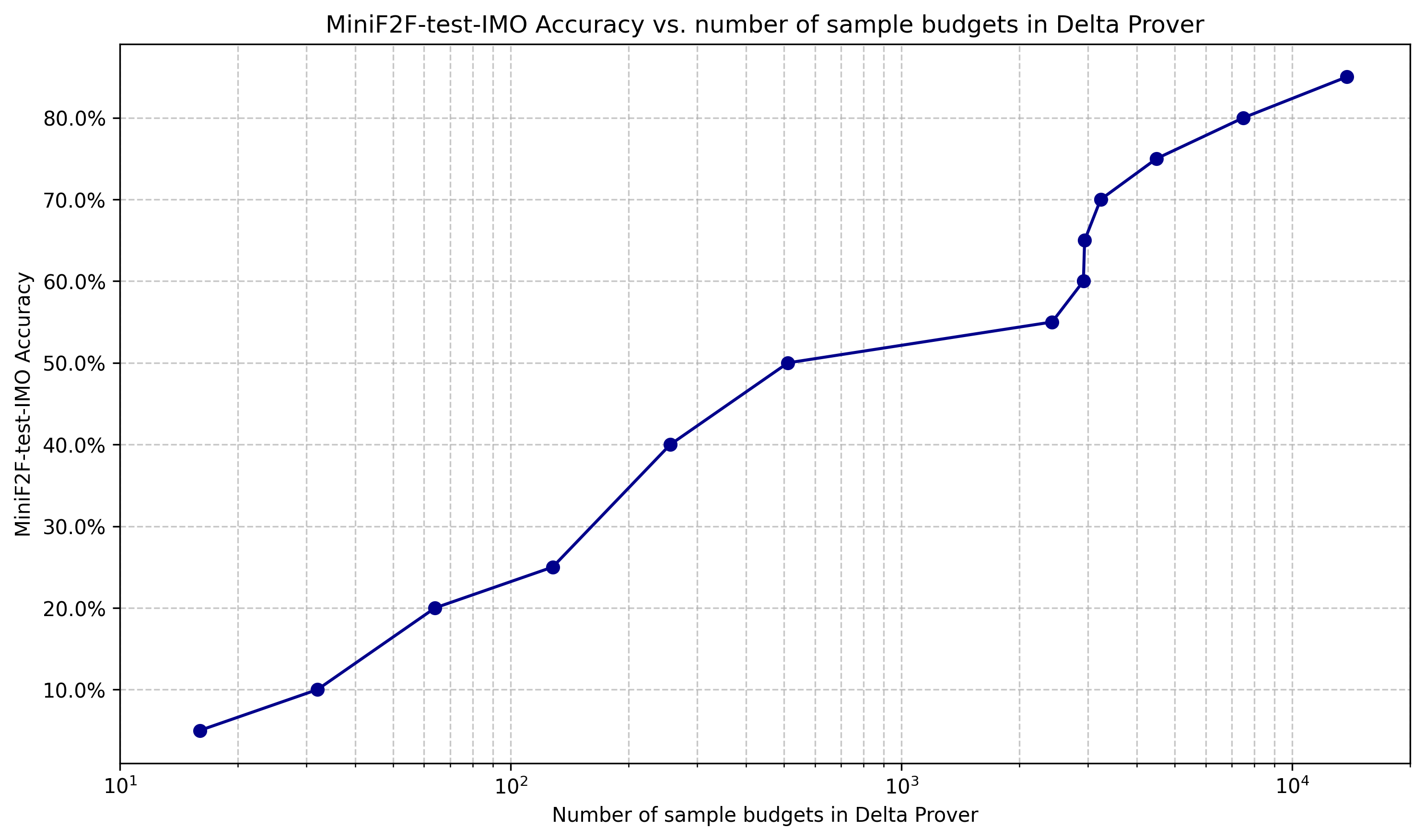} % 替换为你的第二张图片文件名
    \end{subfigure}

    \caption{The test-time scaling law of Delta Prover over the miniF2F-test and the miniF2F-test-IMO benchmarks.} % 整个figure的总标题
    \label{fig:scaling law} % 整个figure的总标签
\end{figure}

Due to the sequential nature of components like proof repair in our approach, the sample budget reported in Table \ref{tab:comparison} is determined as follows: once Delta Prover solves a problem, we record the number of API calls (or equivalent computational budget) used for that successful attempt. The final sample budget is then the \textbf{maximum} of these values across all solved problems. These per-problem statistics also enable us to illustrate Delta Prover's test-time scaling law, as plotted in Figure \ref{fig:scaling law}.

% \begin{figure}[htbp]
%     \centering
%     \includegraphics[width=0.8\textwidth]{sections/figure/solved_cnt.png}  % 替换为你的图片文件名
%     \caption{The test-time scaling law of RAP over the miniF2F-test benchmark.}
%     \label{fig:scaling law}
% \end{figure}

% We deliberately exclude results from existing training-free methods in Table~\ref{tab:comparison}, even though the best reported accuracy among them is 50\% by LEGO-Prover~\cite{wang2023lego}.
% This decision stems from the fact that their results are not directly comparable to ours.
% These earlier methods employed general-purpose Large Language Models (LLMs) like GPT-4---which, while state-of-the-art at the time of their experiments, are superseded in capability by more powerful, recently developed LLMs such as DeepSeek-V3, which our work utilizes.
% Thus, a direct comparison would not accurately reflect the relative advancements in prover performance.

\begin{table}[htbp]
\centering
\caption{Comparison with state-of-the-art approaches on the miniF2F-test dataset. The sample budgets and accuracy of state-of-the-art methods are found in the corresponding papers.}
\label{tab:comparison}
\begin{tabular}{lcrr}
\toprule
Method & Training-free & Sample budget & Accuracy \\
\midrule
\multicolumn{4}{l}{State-of-the-art Methods} \\
Goedel-Prover-SFT \cite{lin2025goedel} & \ding{54} & 25600 & 64.7\% \\
STP \cite{dong2025stp} & \ding{54} & 25600 & 67.6\% \\
HunyuanProver 7B \cite{li2024hunyuanprover} & \ding{54} & 1920000 & 68.4\% \\
BFS-Prover 7B \cite{xin2025bfs} & \ding{54} & 2457600 & 70.8\% \\
DeepSeek-Prover-V2 671B \cite{RenEtAl2025:DeepSeekProverV2} & \ding{54} & $8192$ & 88.9\% \\
Kimina-Prover 72B \cite{kimina_prover_2025_new} & \ding{54} & 42000 & 92.2\% \\
\midrule
Gemini 2.5 Pro 05-06 (BoN) & \ding{52} & 16384 & 49.1\% \\
\midrule
Delta Prover (Our Approach) & \ding{52} & 16384 & 95.9\% \\
% & & 10000 & 93.0\% \\
% \midrule
% \multirow{4}{*}{Kimina-Prover-Preview (Wang et al., 2025)} & \multirow{4}{*}{72B} & 1 & 52.94\% \\
% & & 32 & 68.85\% \\
% & & 1024 & 77.87\% \\
% & & 8192 & 80.74\% \\
% \midrule
% \multirow{8}{*}{DeepSeek-Prover-V2 (non-CoT)} & \multirow{4}{*}{7B} & 1 & 55.5\% $\pm$ 1.4\% \\
% & & 32 & 68.0\% $\pm$ 0.5\% \\
% & & 1024 & 73.2\% $\pm$ 0.5\% \\
% & & 8192 & 75.0\% \\
% \cline{2-4}
% & \multirow{4}{*}{671B} & 1 & 59.5\% $\pm$ 1.4\% \\
% & & 32 & 73.8\% $\pm$ 0.4\% \\
% & & 1024 & 76.7\% $\pm$ 0.2\% \\
% & & 8192 & 78.3\% \\
% \midrule
% \multirow{8}{*}{DeepSeek-Prover-V2 (CoT)} & \multirow{4}{*}{7B} & 1 & 58.6\% $\pm$ 1.1\% \\
% & & 32 & 75.6\% $\pm$ 0.5\% \\
% & & 1024 & 79.9\% $\pm$ 0.3\% \\
% & & 8192 & 82.0\% \\
% \cline{2-4}
% & \multirow{4}{*}{671B} & 1 & 61.9\% $\pm$ 1.6\% \\
% & & 32 & 82.4\% $\pm$ 0.6\% \\
% & & 1024 & 86.6\% $\pm$ 0.3\% \\
% & & 8192 & \textbf{88.9\%} \\
\bottomrule
\end{tabular}
\end{table}

\subsection{Ablation Studies}
\label{sec: ablation}

As detailed in Section \ref{sec:framework}, our prover agent employs two primary components: Proof Generation via Iterative Proof Repair and Reflective Decomposition. Given the strong performance reported previously, it is natural to investigate the extent to which each component contributes to the final results. This section presents ablation studies designed to quantify these individual contributions.

\begin{figure}[htbp]
    \centering % 整体居中，使得两个子图作为一个整体在页面上居中

    % --- 第一张子图 ---
    \begin{subfigure}[b]{0.48\textwidth} % 子图宽度，例如文本宽度的48%
        \centering % 子图内部的图片居中
        \includegraphics[width=\textwidth]{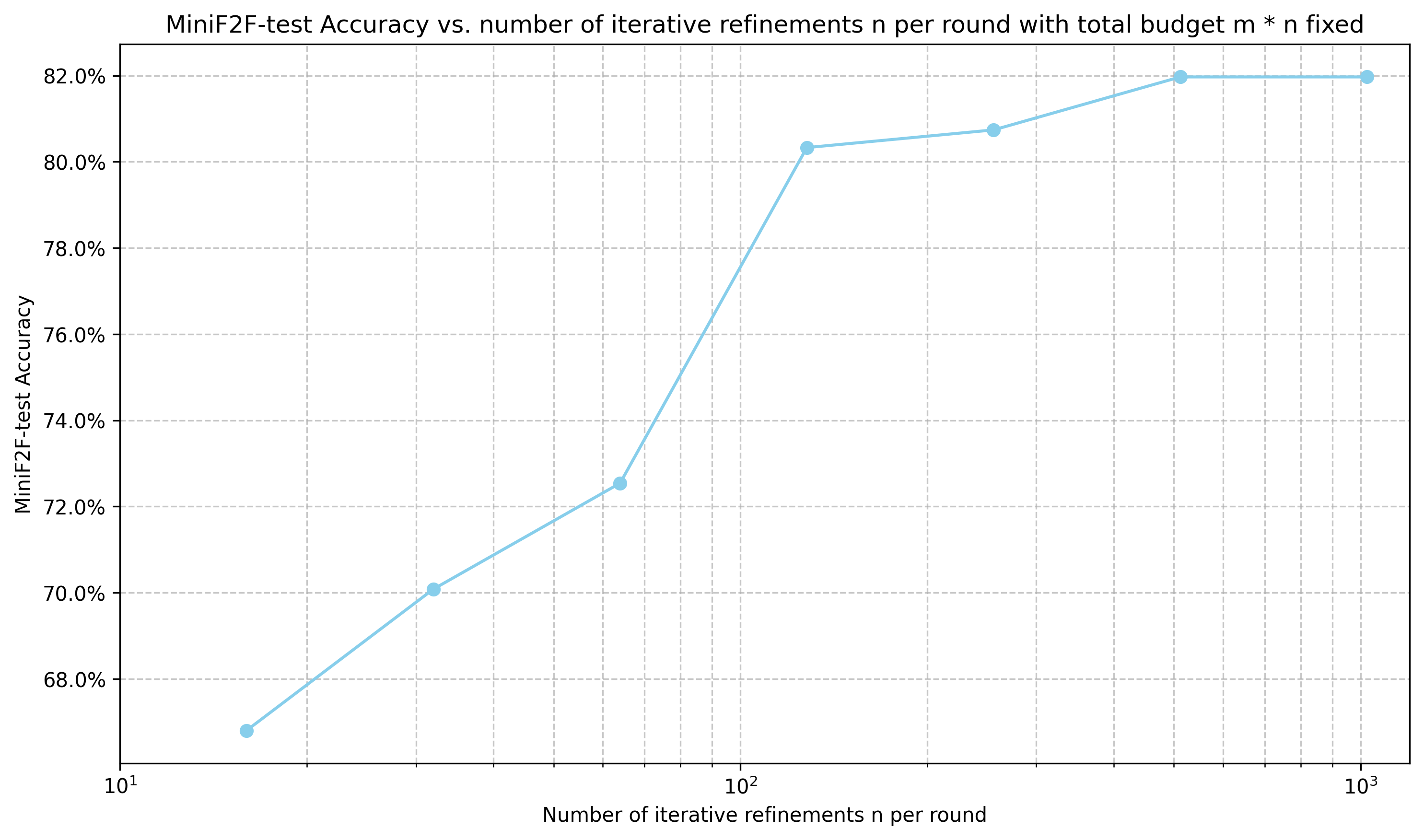} % 替换为你的第一张图片文件名
    \end{subfigure}%  <--- 注意：这里最好有一个 % 来避免潜在的意外空格导
    \hspace{3pt}
    % --- 第二张子图 ---
    \begin{subfigure}[b]{0.48\textwidth} % 子图宽度，例如文本宽度的48%
        \centering % 子图内部的图片居中
        \includegraphics[width=\textwidth]{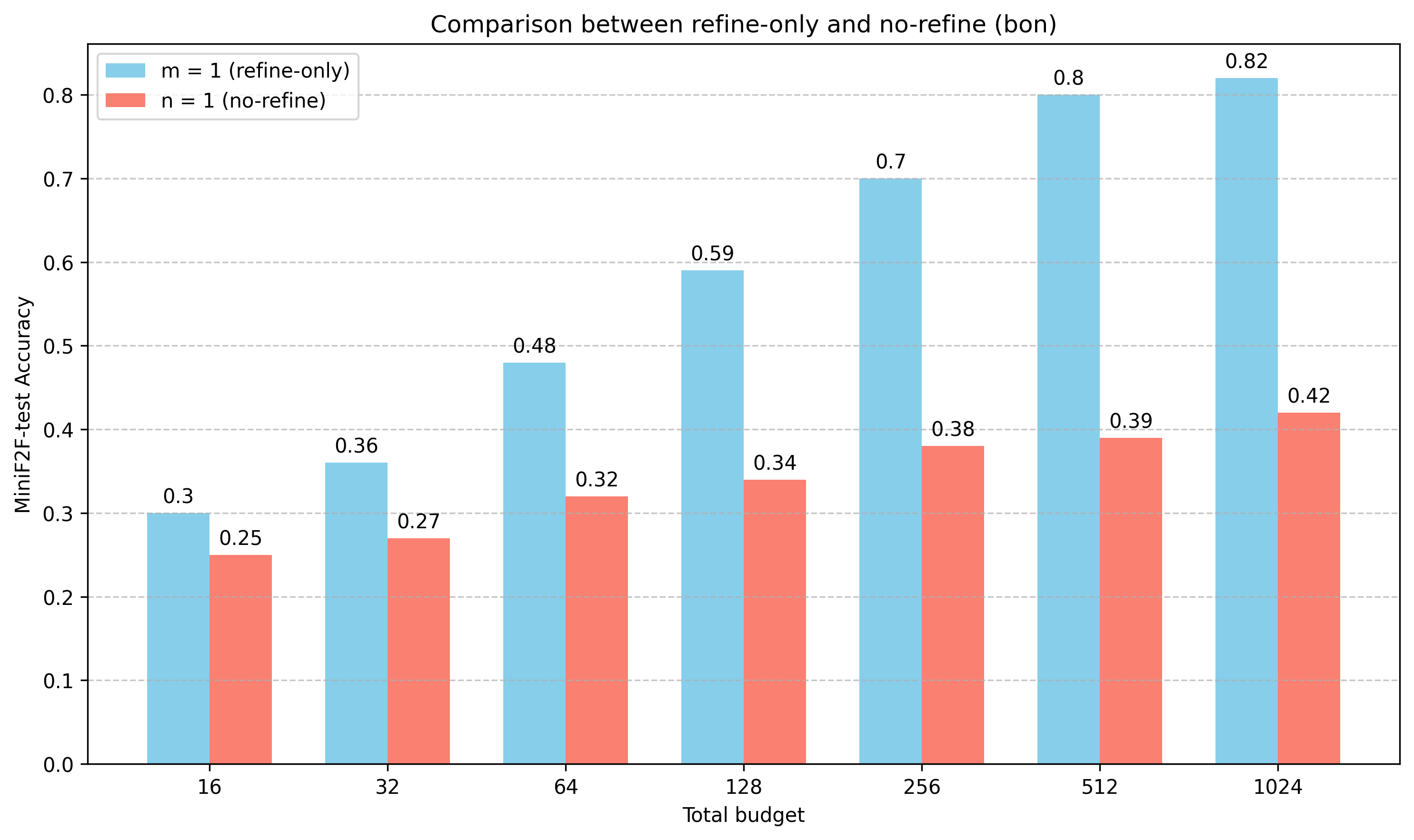} % 替换为你的第二张图片文件名
    \end{subfigure}

    \caption{Ablation study on iterative error correction. Left: We keep the total budget fixed (as 1024), and observe the performance by varying the number of iterative repairs per round (i.e., $n$) and the number of rounds (i.e., $m$). Right: We compare the performance for the repair-only setting (i.e., $m=1$) and no-repair setting (i.e., $n=1$).} % 整个figure的总标题
    \label{fig:ablation_refine} % 整个figure的总标签
\end{figure}

\textbf{Effect of Iterative Proof Repair.} % Or \paragraph{On the effect...}
We first isolate the impact of Iterative Proof Repair (Algorithm \ref{alg:proof_refinement}) by using it exclusively for proof construction. For these experiments, we utilize Gemini 2.5 Pro 05-06 \cite{gemini25pro} as the backbone large language model. Specifically, we conduct the following two experiments:

\begin{itemize}

    \item \textbf{Optimal Hyperparameter Configuration.} We fix the total budget as $1024$ and vary the number of iterative repairs per round $n$ and the number of rounds $m$ to see what is the optimal hyperparameter choice under a prefixed budget. The result can be seen in the left figure of Figure \ref{fig:ablation_refine}. Given that a higher $n$ allows the model more opportunities to fix errors within a proof trajectory, while a higher $m$ explores more distinct initial ideas, one might anticipate a trade-off between $m$ and $n$. However, our results reveal a consistent trend within the tested budget range: accuracy generally improves as the number of repairs $n$ increases for a fixed total budget $m \times n$. This finding, while perhaps counter-intuitive to the exploration-exploitation trade-off, highlights the potent self-correction capabilities inherent in modern large language models.

     \item \textbf{Comparison with Best-of-N (BoN) Sampling.} We further consider two extreme settings, i.e., $m=1$ and $n=1$. When the number of repairs $n=1$, Algorithm \ref{alg:proof_refinement} effectively reduces to the standard Best-of-N (BoN) sampling strategy with a total sample budget of $m$. Our experimental setup thus facilitates a direct comparison between Iterative Proof Repair and BoN. As shown in the right figure of Figure \ref{fig:ablation_refine}, Iterative Proof Repair consistently outperforms BoN sampling, and this performance advantage becomes increasingly pronounced as the total computational budget ($m \times n$) increases, demonstrating a stronger test-time scaling law of Iterative Proof Repair.
\end{itemize}

\textbf{Effect of Reflective Decomposition.} Our second experiment investigates the efficacy of Reflective Decomposition, particularly its role in guiding proof sketch generation. 
We selected the IMO 2019 Problem 1 (\texttt{imo\_2019\_p1}) from the MiniF2F test set. 
This problem, stated below, has proven challenging for automated theorem provers, remaining unsolved by prior methods, including the recent DeepSeek-Prover-V2~\cite{RenEtAl2025:DeepSeekProverV2}. 
This makes it an excellent benchmark for our approach.

% \begin{center}
% \begin{minipage}{0.9\linewidth} % Adjust width as needed
% \begin{minted}{lean}
% theorem imo_2019_p1 (f : ℤ → ℤ) :
%   ((∀ a b, f (2 * a) + (2 * f b) = f (f (a + b))) ↔ (∀ z, f z = 0 \/ ∃ c, ∀ z, f z = 2 * z + c)) := by sorry
% \end{minted}
% \end{minipage}
% \end{center}

\begin{figure}[htbp]
    \centering % 整体居中，使得两个子图作为一个整体在页面上居中

    % --- 第一张子图 ---
    \begin{subfigure}[b]{1.0\textwidth} % 子图宽度，例如文本宽度的48%
        \centering % 子图内部的图片居中
        \includegraphics[width=\textwidth]{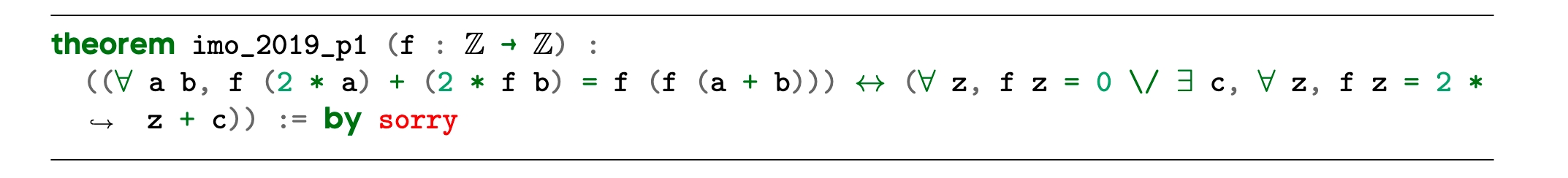} % 替换为你的第一张图片文件名
    \end{subfigure}%  <--- 注意：这里最好有一个 % 来避免潜在的意外空格导
\end{figure}

% \begin{lstlisting}[caption={示例代码}, label={lst:example}]
% …
% \end{lstlisting}

To evaluate the impact of decomposition, we compared two strategies for solving this problem: 
(I) a baseline approach using pure iterative proof correction, and 
(II) Delta Prover, which incorporated iterative proof correction together with Reflective Decomposition.

The baseline approach (pure iterative proof correction) failed to find a solution after 1024 API calls. 
In contrast, Delta Prover successfully solved the problem. The initial Reflective Decomposition round broke the problem into 83 sub-problems. 
Each sub-problem required, on average, 4 API calls to be solved. 
Consequently, the entire problem was solved using approximately $83 \times 4 = 332$ API calls. 
This is significantly fewer than the baseline and well under an initial budget of 400 API calls, clearly demonstrating the power and efficiency of Reflective Decomposition. 
The formal sketch found by our method is included in Appendix~\ref{sec: example_appen}.

\section{Conclusion and Future Work}
In conclusion, while the reasoning capabilities of large language models (LLMs) have shown immense promise, their application to formal theorem proving has been significantly challenged by the difficulty of mastering specialized formal languages and the substantial costs associated with bespoke model training. This work introduced Delta Prover, an agent-based framework designed to overcome these hurdles. By leveraging the inherent reasoning and reflection capabilities of general-purpose LLMs within an interactive Lean 4 environment, Delta Prover successfully orchestrates complex proof construction through reflective decomposition and iterative repair, eliminating the need for task-specific fine-tuning or extensive labeled data. Our findings demonstrate that this approach not only achieves state-of-the-art performance on rigorous benchmarks like miniF2F-test, surpassing even specialized models, but also offers a more scalable and resource-efficient pathway. This agent-centric methodology significantly advances automated theorem proving and offers broader insights into harnessing the sophisticated cognitive capacities of LLMs for complex problem-solving.

% As future work,
% we're interested in further boosting the performance in several possible ways: 
% i) To leverage more elaborate test-time technique such as the evolutionary algorithm as in FunSearch \cite{romera2024mathematical} and AlphaEvolve \cite{deepmindalphaproof},
% other than a predefined workflow.
% ii) To enable RL training with the agentic workflow proposed in this paper.

For future work, we aim to further enhance performance through several promising directions:
\begin{itemize}
    \item Exploring more elaborate test-time techniques, such as evolutionary algorithms (e.g., FunSearch \cite{romera2024mathematical}, AlphaEvolve \cite{deepmindalphaproof}), to move beyond predefined workflows towards dynamic proof search optimization.
    \item Implementing reinforcement learning (RL) based on the agentic workflow proposed in this paper, enabling the agent to continuously refine its proof strategies.
\end{itemize}

\clearpage

\bibliographystyle{plainnat}
\bibliography{main}

\clearpage

\beginappendix

\section{Prompt Templates}
\begin{figure}[htbp]
    \centering
    \fbox{\includegraphics[width=1.0\textwidth]{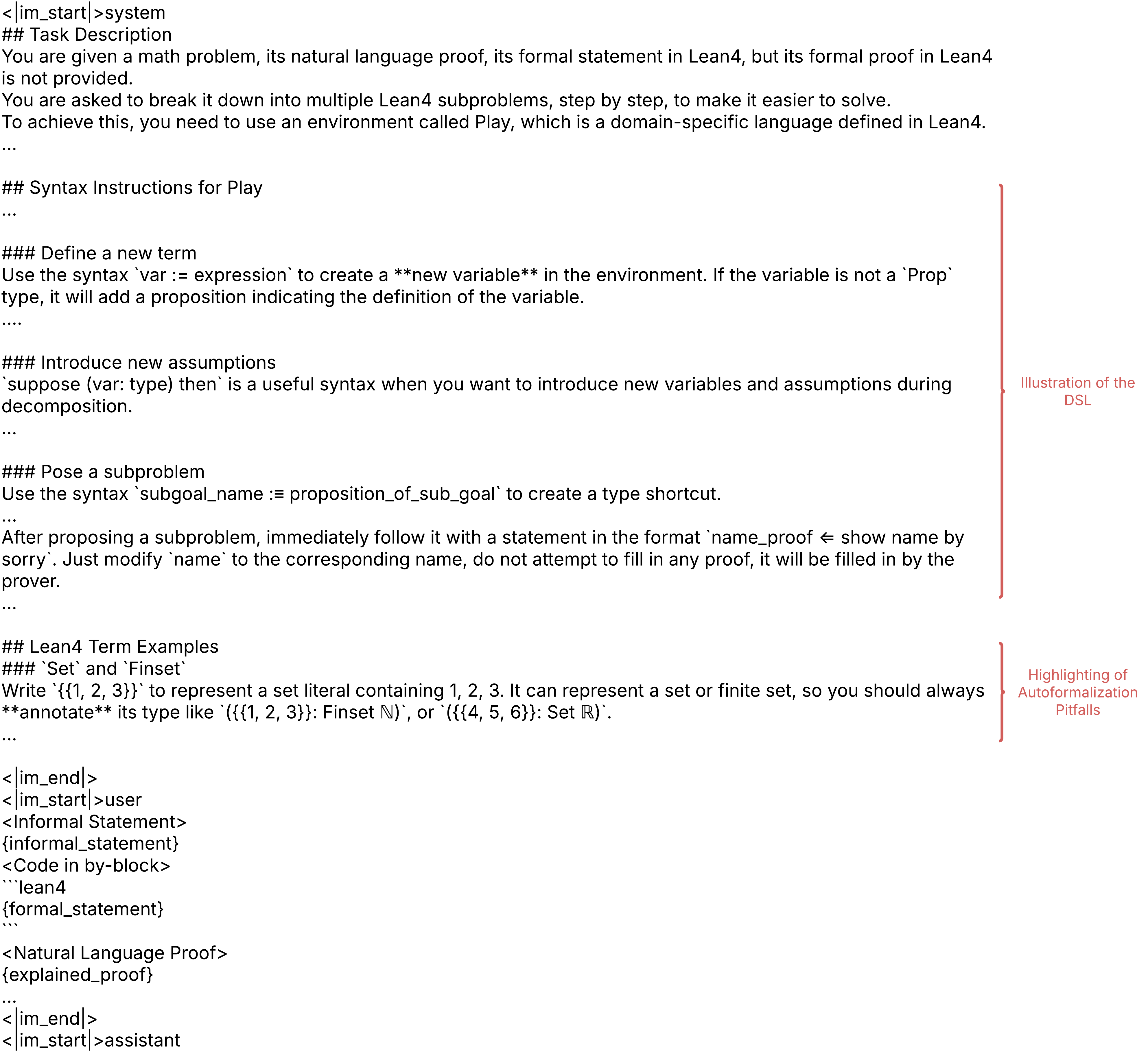}}  % 替换为你的图片文件名
    \caption{The prompt template for sub-problem decomposition.}
    \label{fig:prompt decomp}
\end{figure}

\begin{figure}[htbp]
    \centering
    \fbox{\includegraphics[width=1.0\textwidth]{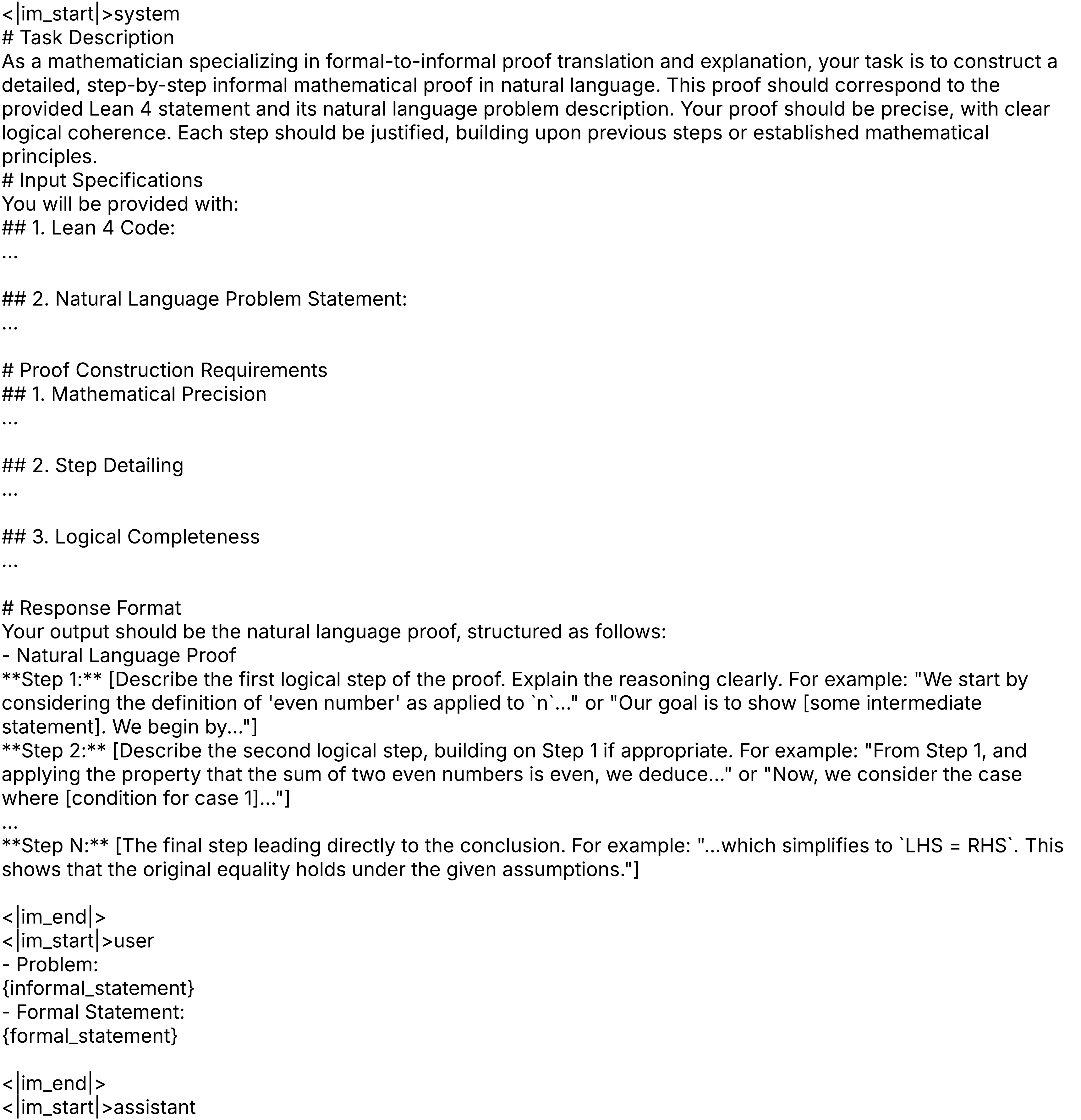}}  % 替换为你的图片文件名
    \caption{The prompt template for natural language sketch.}
    \label{fig:prompt nl}
\end{figure}

\begin{figure}[htbp]
    \centering
    \fbox{\includegraphics[width=1.0\textwidth]{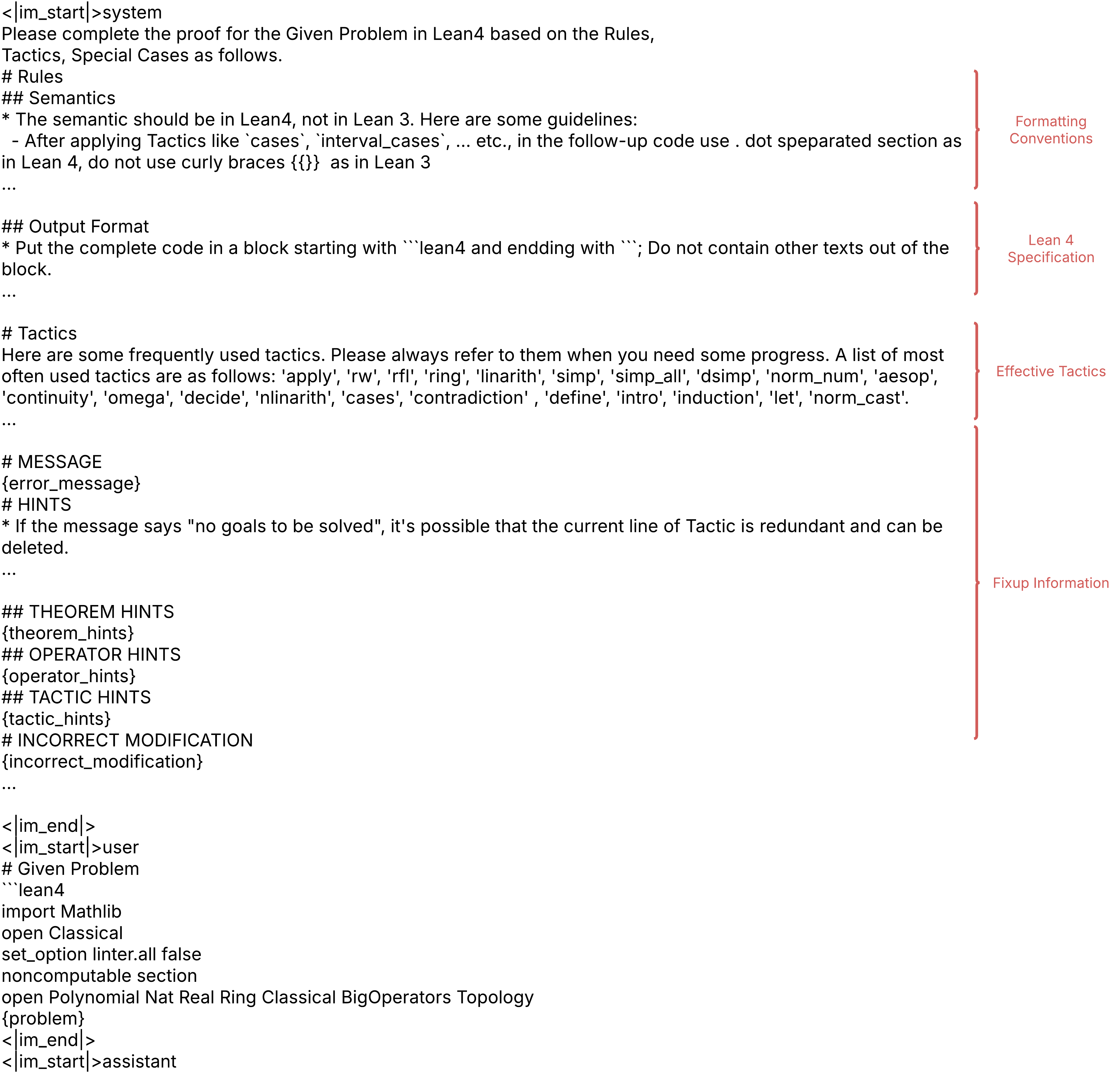}}  % 替换为你的图片文件名
    \caption{The prompt template for proof repair.}
    \label{fig:prompt solve}
\end{figure}

%\subsection{Hello World}
In this section, we provide the concrete templates for the prompt introduced in Section \ref{sec:framework}. Specifically, there are three types of prompts employed in our prover agent, described as follows:
\begin{itemize}
    \item \textbf{Prompt for Sub-problem Decomposition.} As outlined in Section \ref{sec:sketch_generation}, the prompt is composed of the illustration of the DSL and the highlighting of autoformalization pitfalls. We provide the concrete template in Figure \ref{fig:prompt decomp}.

    \item \textbf{Prompt for Informal Proof Generation.} As outlined in Section \ref{sec:sketch_generation}, the prompt is composed of getting refined solutions and scoring solutions at a multi-dimensional fine-grained level. We provide the concrete template in Figure \ref{fig:prompt nl}.
    
    \item \textbf{Prompt for Formal Proof Generation.} As outlined in Section \ref{sec:iterative_error_correction}, the prompt consists of formatting conventions, Lean 4 specification, effective tactics, and fixup information. The corresponding template can be seen in Figure \ref{fig:prompt solve}.
\end{itemize}

% \begin{figure}[htbp]
%     \centering
%     \fbox{\includegraphics[width=1.0\textwidth]{sections/figure/Scoring prompt.pdf}}  % 替换为你的图片文件名
%     \caption{The prompt template for scoring solutions at a multi-dimensional fine-grained level.}
%     \label{fig:prompt score}
% \end{figure}

\newpage
\section{Representative Examples}
\label{sec: example_appen}
% In this section, we provide representative examples illustrating the key components of RAP.

% In the first example, we illustrate how the retrieval mechanism works. In the following Lean code, the LLM generates a wrong theorem name $AntitoneOn.sum\_le\_integral\_Ico$ in its first attempt. Our retrieval mechanism then comes into use, retrieving the correct theorem name and signature based on the wrong one and hinting the LLM.  This highlights how the combination of the LLM's suggestions and our retrieval system enables the discovery and application of advanced mathematical tools that are often difficult to find for humans.

This section presents representative examples illustrating the key components of Delta Prover.

The first example demonstrates our retrieval mechanism. As shown in the accompanying Lean code, the LLM initially generates an incorrect theorem name, \texttt{AntitoneOn.sum\_le\_integral\_Icc}. Our retrieval mechanism then intervenes, using this erroneous suggestion to retrieve the correct theorem name (\texttt{AntitoneOn.sum\_le\_integral\_Ico}) and its signature. This information is subsequently provided as a hint to the LLM. This case highlights how the synergy between the LLM's initial proposals and our retrieval system facilitates the discovery and application of advanced mathematical tools, which can be challenging for humans to find independently.

\begin{figure}[htbp]
    \centering % 整体居中，使得两个子图作为一个整体在页面上居中

    % --- 第一张子图 ---
    \begin{subfigure}[b]{\textwidth} % 子图宽度，例如文本宽度的48%
        \centering % 子图内部的图片居中
        \includegraphics[width=0.8\textwidth]{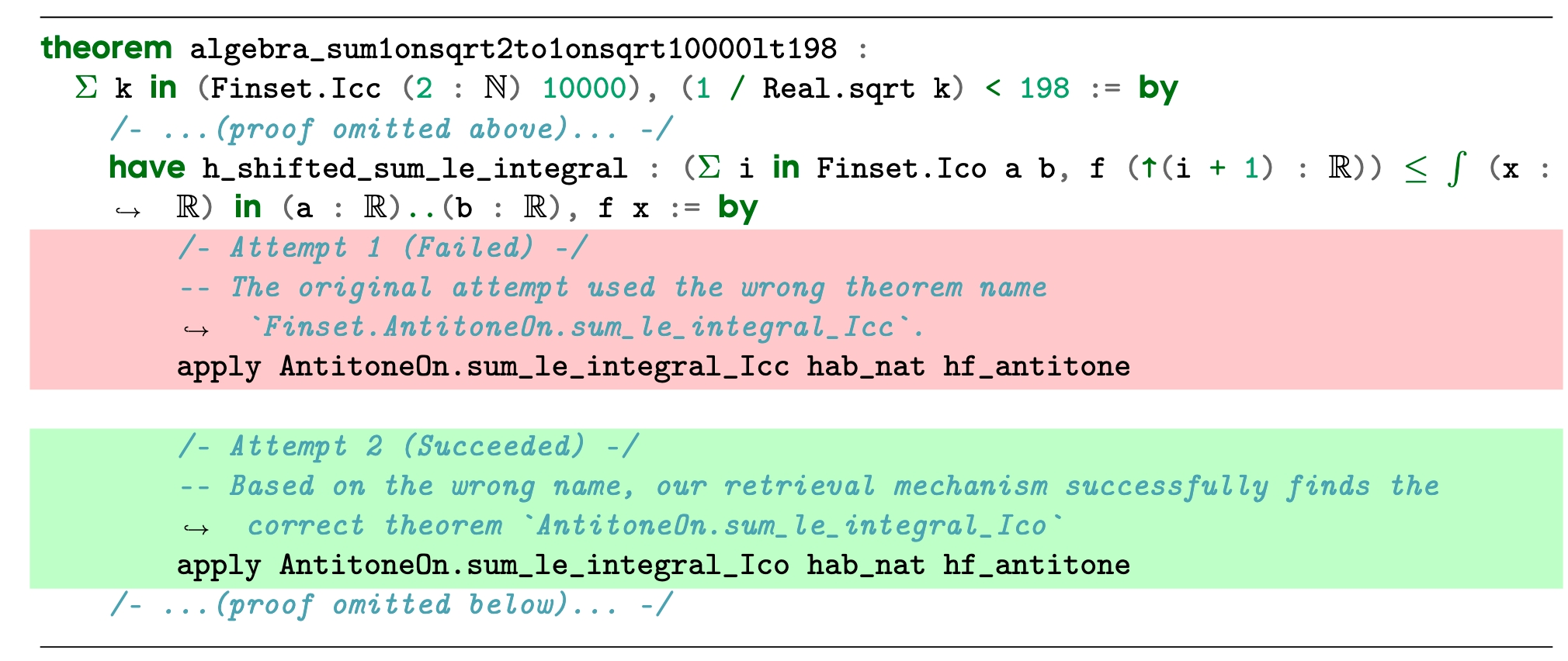} % 替换为你的第一张图片文件名
    \end{subfigure}%  <--- 注意：这里最好有一个 % 来避免潜在的意外空格导
\end{figure}

% \textcolor{blue}{The following Lean code demonstrates iterative proof repairment capability. Although Gemini has not been specifically trained on Lean, it is still able to attempt the use of the theorem $mul\_div\_cancel\_left$ to solve the problem. When this approach leads to an error due to incorrect usage, Gemini is able to reflect on the mistake and identify a simpler solution, such as employing the $simp$ tactic. Furthermore, when Lean returns an error indicating that $simp$ alone is insufficient to complete the arithmetic simplification, Gemini recognizes the limitation and explicitly invokes $norm\_num$ to finalize the proof. This process demonstrates Gemini’s iterative problem-solving and self-correction abilities, as it adapts its approach in response to feedback and progressively refines its proof strategy.}

% The next example further  provides a more complicated case of iterative proof repairment, where the LLM reflects two times before reaching at the correct solution.  The LLM first tries to use the theorem $mul\_div\_cancel\_left$ to solve the problem, but leading to an error. The LLM then  reflect on the mistake and identify an alternative approach, i.e., employing the $simp$ tactic. However, Lean still returns an error indicating that $simp$ alone is insufficient to complete the arithmetic simplification. The LLM recognizes the limitation and explicitly invokes $norm\_num$ to finalize the proof. 

\newpage
Our subsequent example showcases a more intricate scenario of iterative proof correction, requiring the LLM to reflect twice to reach the correct solution. The LLM's first attempt involves using the theorem \texttt{mul\_div\_cancel\_left}, but this results in an error. Following this, the LLM reflects on the failure and proposes an alternative: the \texttt{simp} tactic. When Lean reports another error, signifying that \texttt{simp} alone cannot complete the arithmetic simplification, the LLM again adapts. It recognizes this further limitation and explicitly calls \texttt{norm\_num} to finalize the proof.

\begin{figure}[htbp]
    \centering % 整体居中，使得两个子图作为一个整体在页面上居中

    % --- 第一张子图 ---
    \begin{subfigure}[b]{\textwidth} % 子图宽度，例如文本宽度的48%
        \centering % 子图内部的图片居中
        \includegraphics[width=0.8\textwidth]{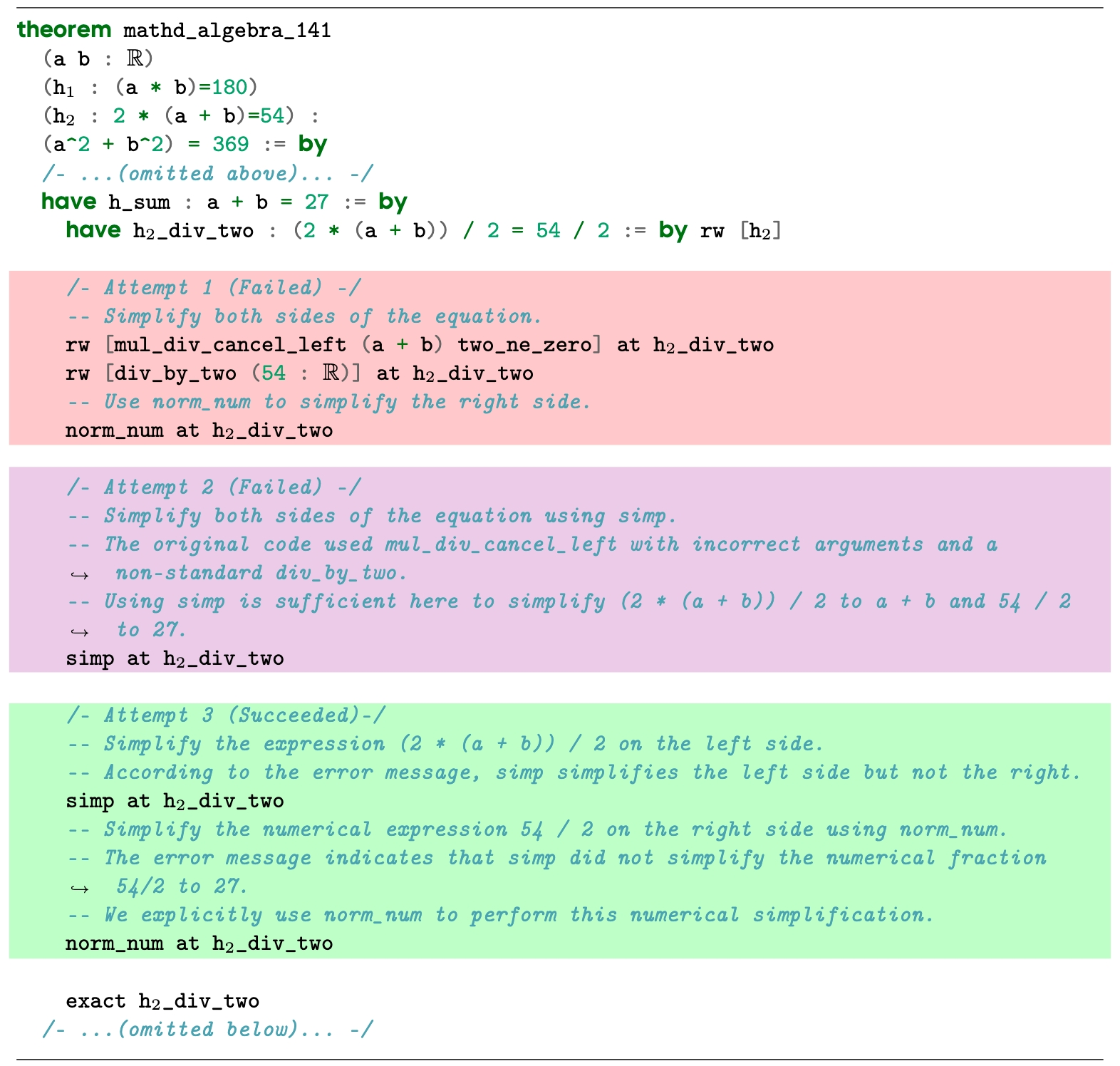} % 替换为你的第一张图片文件名
    \end{subfigure}%  <--- 注意：这里最好有一个 % 来避免潜在的意外空格导
\end{figure}

% The next example showcases the decomposed subproblems of the challenging IMO 2019 P1 problem, as mentioned in the main text. Specifically, the LLM decompose the problem into a very fine grid, i.e., 83 subproblems and facilitates the latter solving process.

\newpage
The next example showcases the LLM's decomposition of the challenging IMO 2019 P1 problem, as referenced in the main text. Specifically, the LLM breaks down the problem into a remarkably fine-grained set of 83 subproblems, thereby facilitating the subsequent solving process.

\begin{figure}[htbp]
    \centering % 整体居中，使得两个子图作为一个整体在页面上居中

    % --- 第一张子图 ---
    \begin{subfigure}[b]{1.0\textwidth} % 子图宽度，例如文本宽度的48%
        \centering % 子图内部的图片居中
        \includegraphics[width=0.8\textwidth]{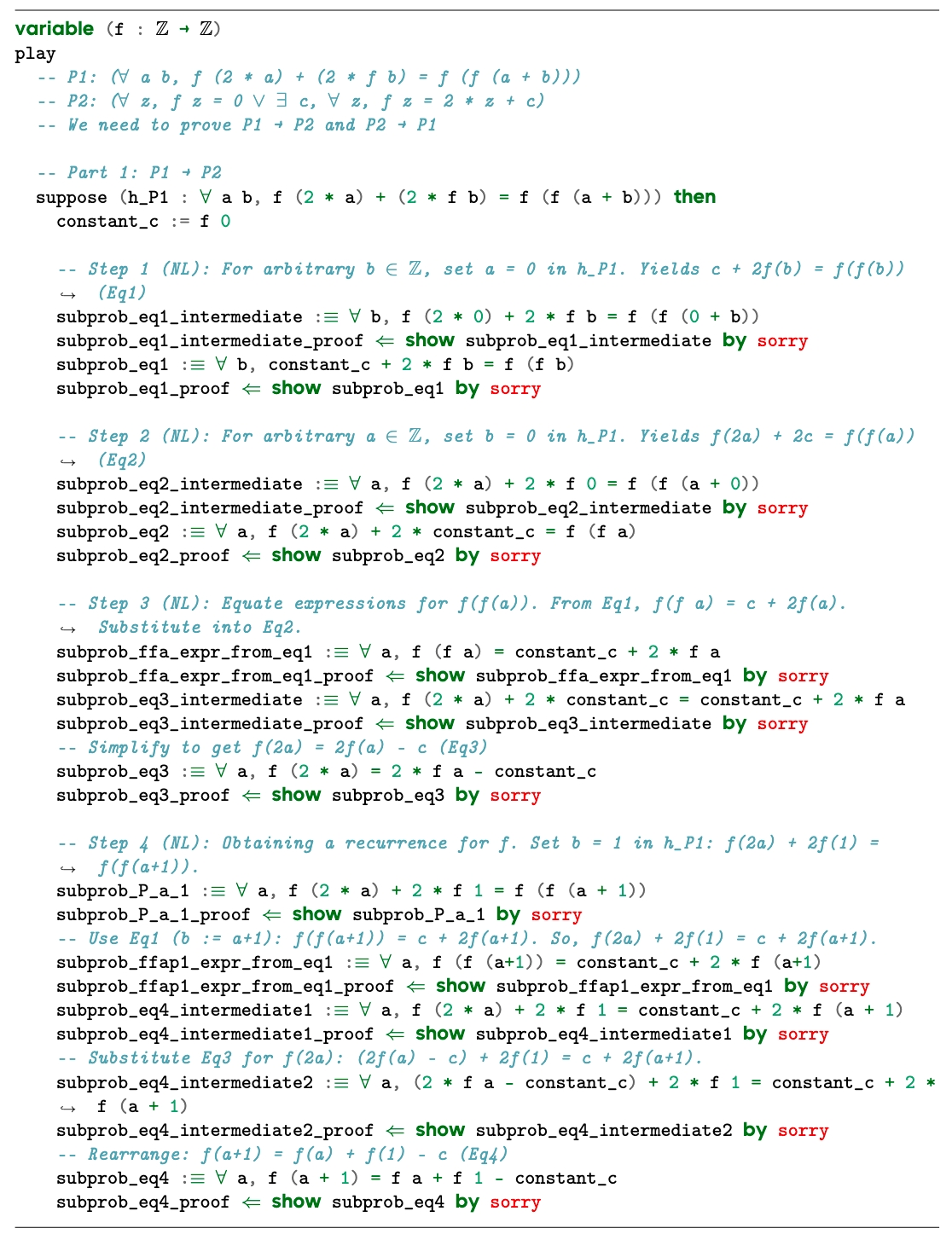} % 替换为你的第一张图片文件名
    \end{subfigure}%  <--- 注意：这里最好有一个 % 来避免潜在的意外空格导
\end{figure}

\begin{figure}[htbp]
    \centering % 整体居中，使得两个子图作为一个整体在页面上居中

    % --- 第一张子图 ---
    \begin{subfigure}[b]{1.0\textwidth} % 子图宽度，例如文本宽度的48%
        \centering % 子图内部的图片居中
        \includegraphics[width=0.8\textwidth]{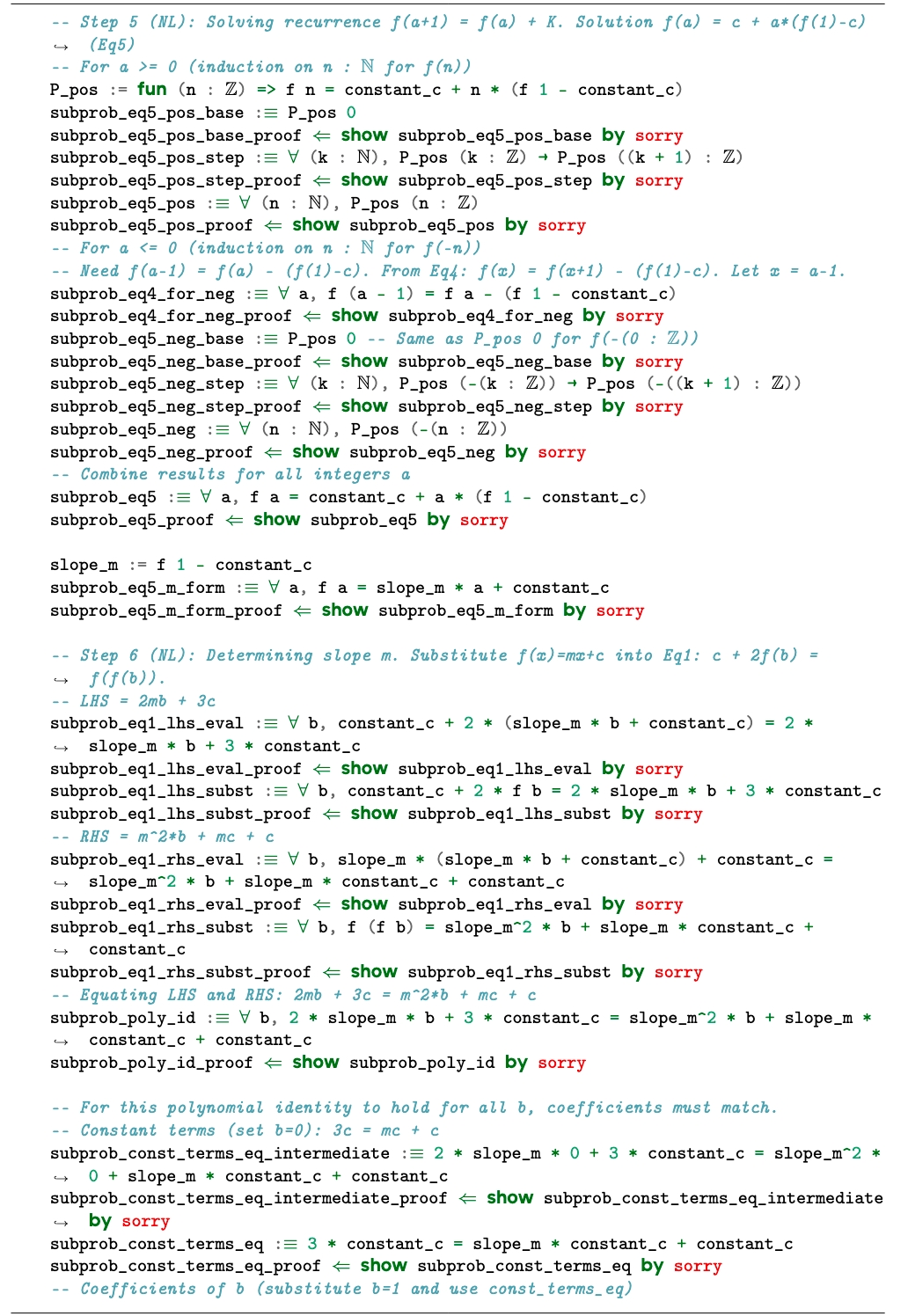} % 替换为你的第一张图片文件名
    \end{subfigure}%  <--- 注意：这里最好有一个 % 来避免潜在的意外空格导
\end{figure}

\begin{figure}[htbp]
    \centering % 整体居中，使得两个子图作为一个整体在页面上居中

    % --- 第一张子图 ---
    \begin{subfigure}[b]{1.0\textwidth} % 子图宽度，例如文本宽度的48%
        \centering % 子图内部的图片居中
        \includegraphics[width=0.8\textwidth]{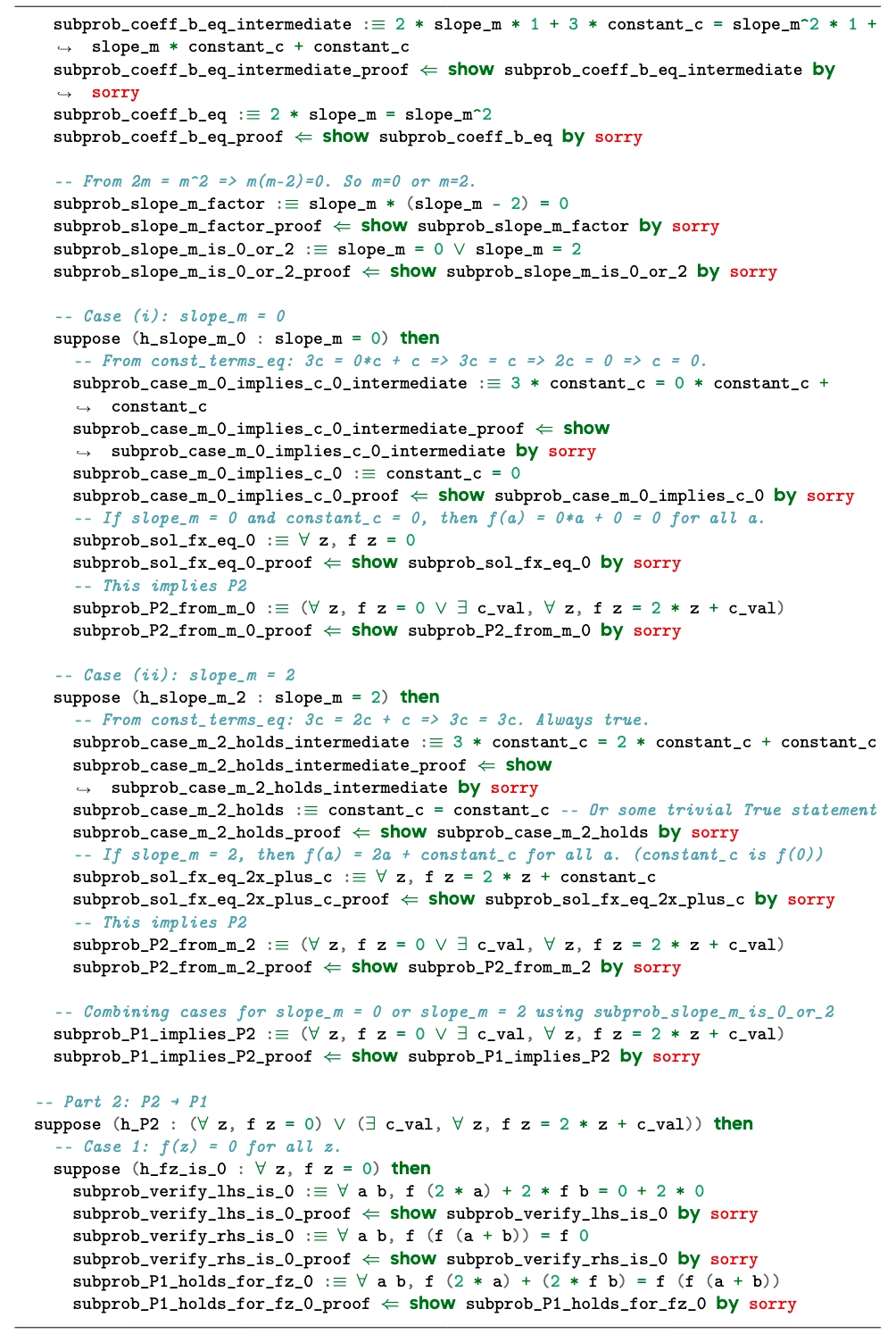} % 替换为你的第一张图片文件名
    \end{subfigure}%  <--- 注意：这里最好有一个 % 来避免潜在的意外空格导
\end{figure}

\begin{figure}[htbp]
    \centering % 整体居中，使得两个子图作为一个整体在页面上居中

    % --- 第一张子图 ---
    \begin{subfigure}[b]{1.0\textwidth} % 子图宽度，例如文本宽度的48%
        \centering % 子图内部的图片居中
        \includegraphics[width=0.8\textwidth]{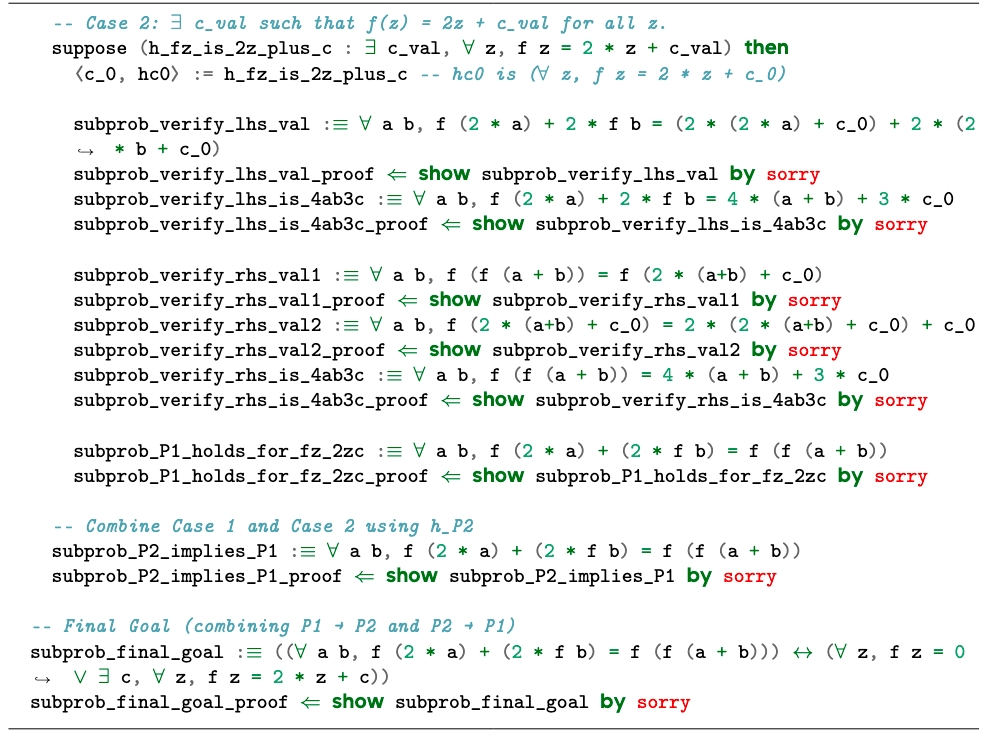} % 替换为你的第一张图片文件名
    \end{subfigure}%  <--- 注意：这里最好有一个 % 来避免潜在的意外空格导
\end{figure}

\end{document}